\documentclass{article}

\usepackage[main, final]{neurips_2026}

\usepackage[utf8]{inputenc} 
\usepackage[T1]{fontenc}    
\usepackage{hyperref}       
\usepackage{url}            
\usepackage{booktabs}       
\usepackage{amsfonts}       
\usepackage{amsmath}        
\usepackage{amsthm}         
\usepackage{nicefrac}       
\usepackage{microtype}      
\usepackage[table]{xcolor}  
\usepackage{listings}       
\usepackage{dsfont}
\usepackage{graphicx}
\usepackage{enumitem}
\usepackage{subcaption}
\definecolor{TableLow}{RGB}{0, 86, 190}
\definecolor{TableHigh}{RGB}{205, 54, 0}
\definecolor{TableStripe}{RGB}{244, 245, 247}
\definecolor{CodeBg}{RGB}{248, 250, 252}
\definecolor{CodeFrame}{RGB}{203, 213, 225}
\definecolor{CodeText}{RGB}{15, 23, 42}
\definecolor{CodeKeyword}{RGB}{37, 99, 235}
\definecolor{CodeString}{RGB}{180, 83, 9}
\definecolor{CodeComment}{RGB}{22, 101, 52}
\definecolor{CodeNumber}{RGB}{100, 116, 139}
\definecolor{CodeFunc}{RGB}{124, 58, 237}
\lstdefinestyle{idepython}{
    language=Python,
    backgroundcolor=\color{CodeBg},
    basicstyle=\ttfamily\footnotesize\color{CodeText},
    keywordstyle=\color{CodeKeyword}\bfseries,
    commentstyle=\color{CodeComment}\itshape,
    stringstyle=\color{CodeString},
    emph={mask_module_forward,mask_module_agp,phi_mlp,phi_up,agp_phi_up,agp_sigmoid,transpose,reshape,sigmoid,full_like,square,einsum},
    emphstyle=\color{CodeFunc},
    numbers=left,
    numberstyle=\scriptsize\color{CodeNumber},
    numbersep=8pt,
    frame=single,
    framerule=0.5pt,
    rulecolor=\color{CodeFrame},
    xleftmargin=1.2em,
    xrightmargin=0.5em,
    framexleftmargin=1.0em,
    framexrightmargin=0.4em,
    framesep=5pt,
    breaklines=true,
    breakatwhitespace=true,
    columns=fullflexible,
    keepspaces=true,
    showstringspaces=false,
    tabsize=4,
    morekeywords={def,return},
}
\newcommand{\scorecell}[2]{\textcolor{TableHigh!#1!TableLow}{#2}}
\newcommand{\bestscore}[2]{\textcolor{TableHigh!#1!TableLow}{\textbf{#2}}}
\newcommand{\secondscore}[2]{\textcolor{TableHigh!#1!TableLow}{\underline{#2}}}
\theoremstyle{definition}
\newtheorem{definition}{Definition}

\title{RiSE: \underline{R}es\underline{i}dual \underline{S}ubspace \underline{E}xpert for Generalizable \\ Text-Centric Image Forgery Localization}

\author{%
  Kahim Wong$^{1}$ \quad Kemou Li$^{1}$ \quad Yiming Chen$^{1}$ \quad Haiwei Wu$^{2}$ \quad Jiantao Zhou$^{1}$\thanks{Corresponding author: Jiantao Zhou (jtzhou@um.edu.mo).} \\
  $^{1}$ State Key Laboratory of Internet of Things for Smart City, University of Macau \\
  $^{2}$ School of Information and Software Engineering, \\ University of Electronic Science and Technology of China  \\
}

\begin{document}

\maketitle

\begin{abstract}
    As AI-assisted image editing becomes increasingly prevalent, Text-Centric Image Forgery Localization (TFL) is essential for protecting trust in financial and legal records. However, we observe that existing TFL detectors predominantly rely on Full-Parameter Fine-Tuning (FPFT) and fail to generalize to unseen forgery types because FPFT can induce low-rank feature collapse and overfit training artifacts. Meanwhile, prior detectors capture diverse forgery traces with a single unified representation, making subtle traces entangled and difficult to learn. To address these challenges, we propose \textbf{RiSE}, a Residual Subspace Expert model that preserves pre-trained priors by freezing principal SVD components and adapting only the residual subspaces with a ViT-only segmentor to avoid feature collapse. RiSE introduces forgery-aware residual experts, where each expert is trained independently on a training subset with a particular forgery type, and we show that cross-domain localization performance improves consistently as more experts are added. At inference, we introduce Latent Perturbation Confidence (LPC), which selects the most confident expert by analytically computing the latent-perturbed output. LPC enables efficient confidence estimation with a single forward pass, avoiding repeated forward passes on perturbed samples and improving expert selection.
    By training on synthetic forgery data, RiSE with 1 expert outperforms prior methods on real-world forgery images by 20.4\% while reducing training steps by $10\times$. Scaling the number of experts to 9 yields a 35.8\% gain. RiSE remains effective with DCT feature, even when the majority of the parameters is frozen with RGB-only pretraining. Code: \url{https://github.com/KahimWong/RiSE}.
\end{abstract}

\section{Introduction}

Text-Centric Image Forgery Localization (TFL) aims to identify manipulated regions in scanned or captured text-centric images. This problem is increasingly important because modern AI-assisted editing tools allow ordinary users to alter document and scene-text images with little expertise \citep{wong2025fontguard, wong2025end, wong2026k},
threatening trust in images used for financial verification, legal review, and identity authentication. Although recent TFL methods have improved accuracy by combining RGB, frequency \citep{qu2023towards}, and multi-scale cues \citep{dong2024robust}, their performance still drops sharply under cross-domain evaluation, where unseen forgery patterns and acquisition artifacts differ from the training distribution \citep{du2025forensichub,ntire2026}. We argue that this failure is not merely a data scarcity issue. Unlike semantic segmentation, TFL depends on weak, non-semantic forensic traces whose statistics vary sharply across editing pipelines, recompression settings, and capture conditions. As a result, many existing TFL detectors \citep{qu2023towards, chen2024enhancing, wong2025adcd} rely on Full-Parameter Fine-Tuning (FPFT) and end up specializing to the limited forgery patterns seen during training. Their penultimate features become highly low-rank, as shown in Fig.~\ref{fig:effective_rank_compare}, indicating representation collapse that harms cross-domain generalization.

This observation points to a different formulation of TFL: the central challenge is to control \emph{where} task-specific adaptation happens. Recent work on residual subspace adaptation \citep{yan2024orthogonal} shows that transferable priors from Vision Foundation Models (VFMs), typically built on the Vision Transformer \citep{dosovitskiy2020image} (ViT) architecture (e.g., CLIP \citep{radford2021learning}, DINOv2 \citep{oquab2024dinov}, and DINOv3 \citep{simeoni2025dinov3}), can be preserved by freezing the dominant SVD components of ViT linear layers and fine-tuning only residual subspaces. This prior-preserving adaptation avoids representation collapse and improves performance in AI-generated image detection \citep{yan2024orthogonal}. However, bringing this idea to TFL requires more than inserting residual adaptation into an off-the-shelf segmentor. In TFL, the task-specific decoders and fusion modules can become a second adaptation pathway, allowing the model to memorize dataset-specific compression, synthesis, or capture artifacts outside the pre-trained backbone. At the same time, learning all forgery types in a single shared residual space entangles heterogeneous traces, causing subtle manipulation cues to be overwhelmed by dominant but less transferable patterns. This leads to two questions: (1) Can TFL be reformulated as a prior-preserving localization problem, with minimal trainable modules beyond the ViT so that adaptation remains constrained and transferable? (2) Can the limited trainable capacity be expertized, so that different residual subspaces specialize to complementary forgery cues and generalize better under domain shift?

To answer these questions, we propose \textbf{RiSE}, a \underline{R}es\underline{i}dual \underline{S}ubspace \underline{E}xpert model for TFL built on a ViT-only segmenter. We instantiate RiSE on EoMT \citep{kerssies2025your}, which turns a ViT into a segmenter using only query embeddings and a lightweight mask module, thereby minimizing additional adaptation pathways beyond the backbone. On top of this architecture, RiSE divides the training set into non-overlapping subsets with a particular forgery type and trains one residual expert on each subset independently. All experts share the dominant principal subspaces of the VFM backbone, while each expert learns only a tiny residual subspace tailored to a more homogeneous forgery regime. We show that this offline image-level routing is more stable than learnable routing \citep{shazeer2017outrageously}, and that scaling the number of experts consistently improves cross-domain performance as shown in Fig.~\ref{fig:expert_scaling} while parameter growth remains sublinear because the principal components are shared among experts. To select the final prediction among experts, we further propose Latent Perturbation Confidence (LPC), which addresses the limitation that existing confidence metrics \citep{hendrycks2017baseline, borges2024soft} often become unreliable when forged regions are small. LPC estimates the expected latent-perturbed output analytically and favors experts whose forged predictions remain stable under perturbations. As a result, LPC enables single-pass confidence estimation, avoiding repeated forward passes on perturbed samples and improving expert selection for more accurate localization. Trained on DocTamper \citep{qu2023towards} synthetic forgery data, a single RiSE expert generalizes effectively to real-world forgery images, achieving a substantial 20.4\% performance improvement while reducing training steps by $10\times$ compared with existing TFL methods. Scaling to 9 experts further boosts the performance gain to 35.8\%.

Our contributions are fourfold:
\begin{itemize}[leftmargin=*,itemsep=0.00em,topsep=0.08em]
    \item We identify uncontrolled task-specific adaptation as a key cause of poor cross-domain TFL, show that FPFT induces low-rank feature collapse, and reformulate TFL as a prior-preserving localization problem with a ViT-only residual adaptation architecture.
    \item We introduce residual subspace experts for TFL, each optimized on a training subset with a specific forgery type, so that different residual subspaces specialize to complementary forgery cues and improve cross-domain localization as more experts are added while keeping parameter growth sublinear.
    \item We develop a plug-and-play confidence estimator, LPC, that analytically computes expected latent-perturbed outputs to measure expert confidence effectively and efficiently.
    \item Extensive experiments demonstrate that RiSE consistently improves cross-domain localization performance by 35.8\% across 4 test sets.
\end{itemize}

\section{Related Work}

\textbf{General image forgery localization} increasingly focuses on robust trace modeling across manipulation types. Representative methods such as MVSS-Net \citep{chen2021image} and PSCC-Net \citep{liu2022pscc} use multi-scale cues, CAT-Net \citep{kwon2022learning} exploits JPEG DCT statistics, and later work including IFOSN \citep{wu2022robust}, TruFor \citep{guillaro2023trufor}, FOCAL \citep{wu2025rethinking}, and Mesorch \citep{zhu2025mesoscopic} improves generalization through modeling Online Social Networks (OSN) transmission, noise fingerprint, contrastive feature, and designing CNN--Transformer structure. Nevertheless, general image forgery detectors often struggle on text-centric images because forged regions can blend into the uniform backgrounds and structured text layouts.

\textbf{Text-centric image forgery localization} has progressed from early benchmarking to DCT-aware and robustness-oriented detectors~\citep{seed}. DTD \citep{qu2023towards} introduced the DocTamper benchmarking and showed that DCT perception with multi-scale decoding can effectively localize tampered regions. FFDN \citep{chen2024enhancing} and ADCD-Net \citep{wong2025adcd} further improved localization through RGB--DCT fusion, with ADCD-Net explicitly addressing DCT misalignment and text-background disparity. TIFDM \citep{dong2024robust} strengthens trace enhancement and multi-scale aggregation under degradations, CAFTB \citep{song2025cross} combines spatial and noise-domain cues with cross-attention fusion, and $\text{DITL}^2$ improves performance through learning from both general and text-centric image forgeries \citep{li2025ditl2}. However, most existing TFL detectors rely on FPFT, which overfits training forgery patterns and induces low-rank feature collapse, harming cross-domain generalization. In contrast, RiSE preserves transferable priors via residual subspace adaptation and uses forgery-aware residual experts for stronger specialization and generalization.

\section{Method}

\begin{figure}[t]
    \centering
    \includegraphics[width=0.85\linewidth]{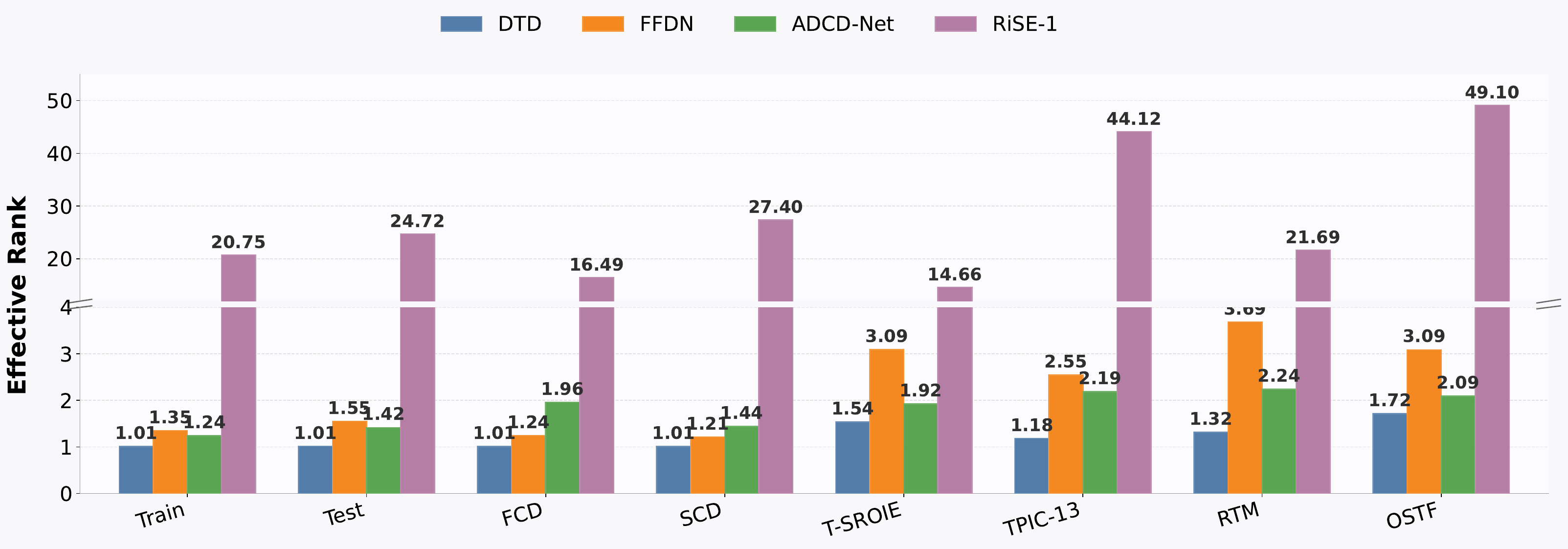}
    \caption{Effective-rank comparison of penultimate features across 8 datasets. Prior TFL detectors (DTD, FFDN, and ADCD-Net) concentrate most representation variance in a very small number of directions, while RiSE-1 (one expert RiSE) maintains a substantially higher effective rank, indicating a less collapsed and more expressive representation.}
    \label{fig:effective_rank_compare}
\end{figure}

\begin{figure}[t]
    \centering
    \begin{subfigure}[t]{0.57\linewidth}
        \centering
        \includegraphics[width=\linewidth]{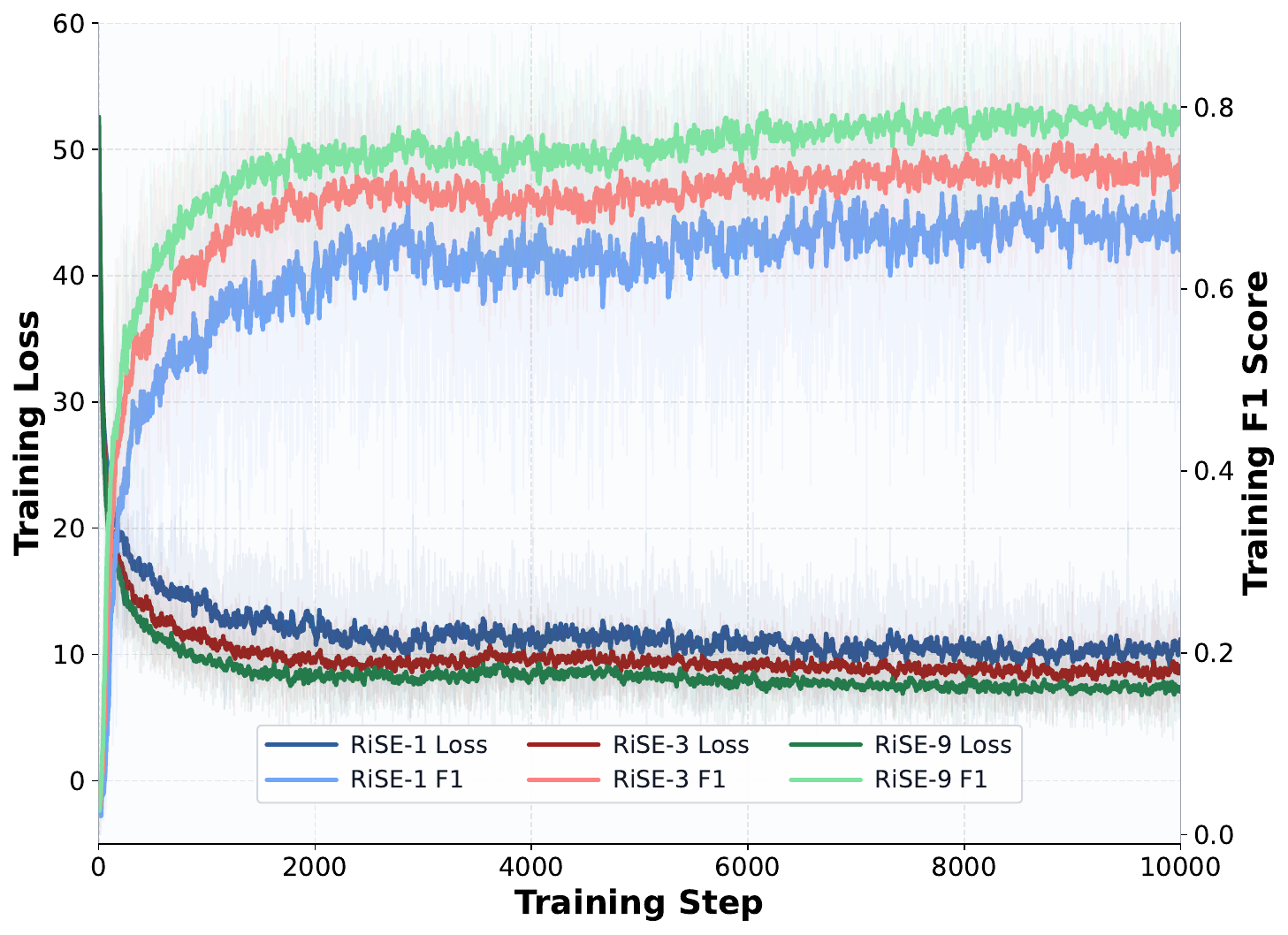}
        \caption{}
        \label{fig:expert_scaling_train_curves}
    \end{subfigure}
    \hfill
    \begin{subfigure}[t]{0.37\linewidth}
        \centering
        \includegraphics[width=\linewidth]{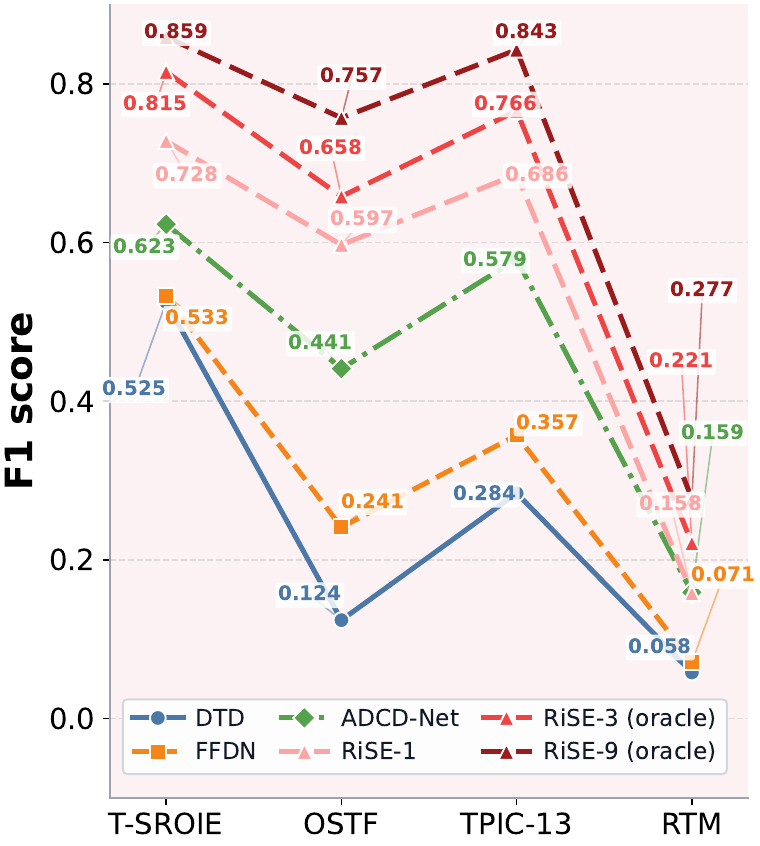}
        \caption{}
        \label{fig:expert_scaling_cross_domain}
    \end{subfigure}
    \caption{More residual experts improve (a) training optimization and (b) cross-domain localization. RiSE-$x$ denotes the RiSE with $x$ residual experts. In (b), RiSE-3 and RiSE-9 report oracle expert selection, i.e., the upper bound obtained by choosing the best expert with the ground-truth mask.}
    \label{fig:expert_scaling}
\end{figure}

\paragraph{Motivation.}

Recent TFL methods such as DTD \citep{qu2023towards}, FFDN \citep{chen2024enhancing}, and ADCD-Net \citep{wong2025adcd} rely on FPFT. As shown in Fig.~\ref{fig:effective_rank_compare}, these models produce strongly low-rank features, overfit to training forgery traces, and perform worse on cross-domain data \citep{du2025forensichub}. Motivated by residual subspace adaptation \citep{yan2024orthogonal}, we use a ViT-only segmenter \citep{kerssies2025your} to keep the architecture minimal and preserve transferable VFM priors. RiSE maintains an effective rank that is roughly $4.7 - 37.3\times$ higher than these prior methods across the 8 datasets, suggesting that residual subspace adaptation better preserves transferable knowledge and avoids feature collapse. The effective rank computation is detailed in Appendix~\ref{app:effective_rank}. Our second motivation is the scaling behavior of residual experts. For TFL, we argue that modeling diverse forgery patterns and image context with a single space could lead to entangled and noisy representations~\citep{featdistill}. Therefore, we propose to split the training data into subsets with a particular forgery type and train a residual expert on each subset independently. Adding residual experts consistently improves training optimization in Fig.~\ref{fig:expert_scaling}~(a), as well as enhancing cross-domain localization in Fig.~\ref{fig:expert_scaling}~(b). For RiSE with multiple experts, we report the oracle performance, which assumes access to the ground-truth mask to choose the best expert for each sample and therefore serves as an upper bound on expert selection. 

\begin{figure}
    \centering
    \includegraphics[width=0.95\linewidth]{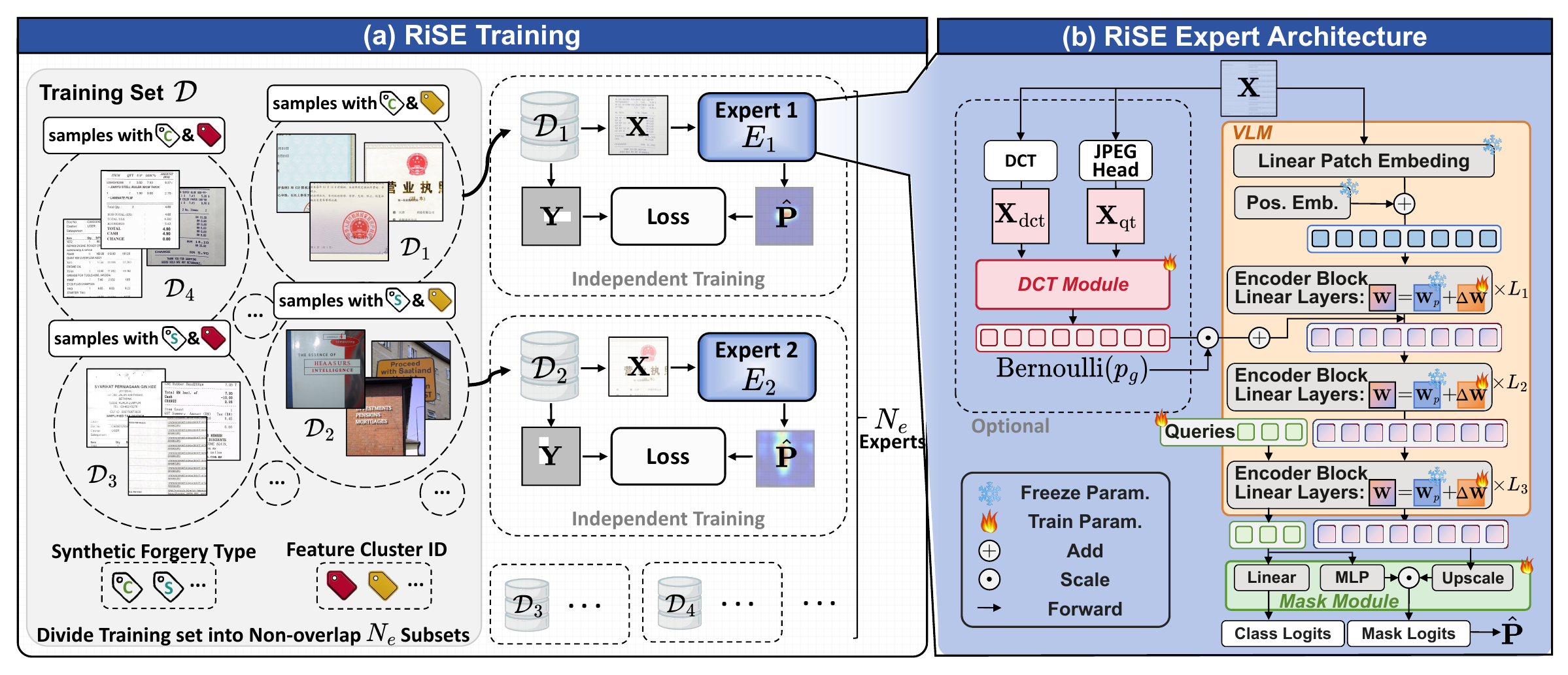}
    \caption{Training overview of RiSE. (a) The training set is partitioned into non-overlapping subsets using forgery synthesis type and feature-cluster index, and each subset trains one expert independently. (b) Each expert uses the same ViT-only segmenter with a shared, optional DCT branch, a shared frozen principal subspace, and expert-specific residual parameters.}
    \label{fig:overview}
\end{figure}

\paragraph{Overview.}
RiSE is a residual subspace expert model for TFL built on a ViT-only segmenter. Let the training set be $\mathcal{D}=\{(\mathbf{X}_i,\mathbf{Y}_i,d_i)\}_{i=1}^{N}$, where $\mathbf{X}_i\in\mathbb{R}^{3\times H\times W}$ is a forgery image, $\mathbf{Y}_i\in\{0,1\}^{H\times W}$ is the ground-truth mask, and $d_i\in\{1,\dots,N_e\}$ is the expert-assignment label of the $i$-th sample. We denote $s_i\in\{1,\dots,N_s\}$ and $k_i\in\{1,\dots,N_c\}$ as the synthesis-type index and feature-cluster index of sample $i$, respectively. We derive the expert-assignment label by $ d_i=(s_i-1)N_c+k_i$. As illustrated in Fig.~\ref{fig:overview}~(a), RiSE partitions the training set by the Cartesian product of $N_s$ forgery synthesis types and $N_c$ feature-cluster indices, yielding $N_e=N_s\times N_c$ non-overlapping sub-training sets $\mathcal{D}=\bigcup_{j=1}^{N_e}\mathcal{D}_j$. Details of the expert assignment are provided in Appendix~\ref{app:domain_label_acquisition}.

\begin{definition}[RiSE Expert]
For expert index $j\in\{1,\dots,N_e\}$, define the $j$-th sub-training set as
\[
\mathcal{D}_j:=\{(\mathbf{X}_i,\mathbf{Y}_i,d_i)\in\mathcal{D}: d_i=j\}.
\]
Let $\Theta_{\mathrm{sh}}$ denote the shared parameter set and let $\Theta_{\mathrm{sp}}^{(j)}$ denote the expert-specific parameter set of expert $j$. Then the $j$-th RiSE expert is the segmentation model
\[
E_j(\cdot;\Theta_{\mathrm{sh}},\Theta_{\mathrm{sp}}^{(j)}):\mathbb{R}^{3\times H\times W}\to[0,1]^{H\times W},
\]
which is optimized only on $\mathcal{D}_j$, i.e.,
\[
(\Theta_{\mathrm{sp}}^{(j)})^\star\in\arg\min_{\Theta_{\mathrm{sp}}^{(j)}}\;\mathbb{E}_{(\mathbf{X},\mathbf{Y},d)\sim\mathcal{D}_j}
\bigl[\mathcal{L}(E_j(\mathbf{X};\Theta_{\mathrm{sh}},\Theta_{\mathrm{sp}}^{(j)}),\mathbf{Y})\bigr].
\]
\end{definition}
In practice, $\Theta_{\mathrm{sp}}^{(j)}$ corresponds to the residual subspace parameters, while $\Theta_{\mathrm{sh}}$ contains the dominant shared pre-trained components and other shared modules. The RiSE expert differs significantly from a conventional Mixture-of-Experts (MoE) \citep{shazeer2017outrageously, kong2025moe}, where the entire model is jointly trained and combined by a learned router. RiSE instead assigns samples to experts offline during training and selects one expert prediction at test time. The detailed architecture of a single expert is shown in Fig.~\ref{fig:overview}~(b). All experts share the same macro architecture while having expert-specific residual parameters in the ViT linear layers. At inference, as shown in Fig.~\ref{fig:infer}~(a), all experts perform inference on the same input in parallel, producing expert-wise probability maps $\{\hat{\mathbf{P}}_j\}_{j=1}^{N_e}$ with $\hat{\mathbf{P}}_j\in[0,1]^{H\times W}$. RiSE then estimates a confidence score $c_j$ for each expert and selects the prediction of the most confident expert, $j^*=\arg\max_j c_j$. 

\subsection{RiSE Architecture and Training}

RiSE partitions the training set into subsets with a particular forgery type, trains experts independently, and selects the most confident expert at inference. We now describe the ViT-only architecture and training details of a single RiSE expert. As shown in Fig.~\ref{fig:overview}~(b), an expert consists of three modules: a ViT-based VFM backbone, an optional DCT module, and a mask module. Given an RGB image $\mathbf{X}$, let $P$ denote the ViT patch size, and $H_p=H/P$ and $W_p=W/P$. RiSE first extracts visual tokens $\mathbf{Z}_\mathrm{rgb}\in\mathbb{R}^{H_pW_p\times d}$ from the first $L_1$ blocks of the ViT. Optionally, a DCT branch extracts JPEG compression traces. The branch first applies an $8\times 8$ block DCT to the luminance channel of $\mathbf{X}$ to obtain coefficients $\mathbf{X}_\mathrm{dct}\in\mathbb{N}^{H\times W}$ and the quantization table $\mathbf{X}_\mathrm{qt}\in\mathbb{N}^{8\times 8}$ from the JPEG header, encodes $(\mathbf{X}_\mathrm{dct}, \mathbf{X}_\mathrm{qt})$ into frequency-aware tokens $\mathbf{Z}_\mathrm{dct}\in\mathbb{R}^{H_pW_p\times d}$, and fuses $\mathbf{Z}_\mathrm{dct}$ with $\mathbf{Z}_\mathrm{rgb}$ to form $\mathbf{Z}_\mathrm{fuse}\in\mathbb{R}^{H_pW_p\times d}$ before the intermediate $L_2$ blocks. The learnable query tokens $\mathbf{Q}\in\mathbb{R}^{N_q\times d}$ are prepended before the last $L_3$ blocks, so that the last VFM block produces both dense tokens $\mathbf{Z}_\mathrm{vfm}\in\mathbb{R}^{H_pW_p\times d}$ and query embeddings $\mathbf{Q}_\mathrm{vfm}\in\mathbb{R}^{N_q\times d}$. The mask module takes $(\mathbf{Z}_\mathrm{vfm}, \mathbf{Q}_\mathrm{vfm})$ and predicts the forgery probability map $\hat{\mathbf{P}}\in[0,1]^{H\times W}$. All experts share the architecture in Fig.~\ref{fig:overview}~(b) and differ only in the expert-specific residual parameters inserted into selected ViT linear layers. Table~\ref{tab:module_id} summarizes the training and sharing status of the modules.

\noindent
\begin{minipage}[t]{0.50\linewidth}
    \vspace{0pt}
    \centering
    \refstepcounter{table}
    {\footnotesize\textbf{Table \thetable.} Module details of a single RiSE expert. ``Shared'' indicates the module is shared by all experts.\par}
    \label{tab:module_id}
    \setlength{\tabcolsep}{3pt}
    \resizebox{\linewidth}{!}{%
    \begin{tabular}{llccc}
        \toprule
        Module & Sub-module & Trainable & Shared & \#Params (M) \\
        \midrule
        DCT Module & - & \checkmark & \checkmark & 3.82 \\
        \midrule
        VFM & Patch Embedding & & \checkmark & 0.79 \\
        & Prefix Tokens / RoPE & & \checkmark & 0.01 \\
        & Principal Weights ($\mathbf{W}_{n-r}$) & & \checkmark & 301.99 \\
        & Residual Weights ($\Delta \mathbf{W}$) & \checkmark & & 0.44 / expert \\
        \midrule
        Query & - & \checkmark & \checkmark & 0.01 \\
        \midrule
        Mask Module & - & \checkmark & \checkmark & 11.56 \\
        \bottomrule
    \end{tabular}%
    }
\end{minipage}\hfill
\begin{minipage}[t]{0.48\linewidth}
    \vspace{0pt}
    \paragraph{VFM backbone.}
    The VFM encodes the input image into visual tokens and processes them with ViT blocks. Residual adaptation applies to any linear layer $\mathbf{W}\in\mathbb{R}^{d_1\times d_2}$. We consider the square case $d_1=d_2=n$, the attention $\texttt{q}/\texttt{k}/\texttt{v}/\texttt{o}$ and MLP $\texttt{up}/\texttt{down}$ linear projections. Given a pre-trained weight matrix $\mathbf{W}\in\mathbb{R}^{n\times n}$, RiSE applies SVD decomposition $\mathbf{W}=\mathbf{U}\mathbf{\Sigma}\mathbf{V}^\top$, where $\mathbf{U}\in\mathbb{R}^{n\times n}$ and $\mathbf{V}\in\mathbb{R}^{n\times n}$ are orthogonal
\end{minipage}

matrices of left and right singular vectors, respectively, and $\mathbf{\Sigma}\in\mathbb{R}^{n\times n}$ is diagonal. We keep the rank-$(n-r)$ approximation $\mathbf{W}_{n-r}=\mathbf{U}_{n-r}\mathbf{\Sigma}_{n-r}\mathbf{V}_{n-r}^\top$, with $\mathbf{U}_{n-r}\in\mathbb{R}^{n\times(n-r)}$, $\mathbf{\Sigma}_{n-r}\in\mathbb{R}^{(n-r)\times(n-r)}$, and $\mathbf{V}_{n-r}\in\mathbb{R}^{n\times(n-r)}$, as the frozen principal components. Each expert shares the same frozen $\mathbf{W}_{n-r}$ and learns an individual residual component $\Delta \mathbf{W}=\mathbf{W}-\mathbf{W}_{n-r}=\mathbf{U}_r\mathbf{\Sigma}_r\mathbf{V}_r^\top$, where $\mathbf{U}_r\in\mathbb{R}^{n\times r}$, $\mathbf{\Sigma}_r\in\mathbb{R}^{r\times r}$, and $\mathbf{V}_r\in\mathbb{R}^{n\times r}$. This residual adaptation preserves the dominant pre-trained knowledge while letting each expert learn specific forgery traces, and the number of parameters grows sublinearly as more experts are added. Before the last $L_3$ encoder blocks, we prepend learnable query tokens $\mathbf{Q}$ so that the backbone produces both query embeddings and feature tokens for downstream localization.

\paragraph{DCT module.}
As demonstrated in prior TFL studies \citep{chen2024enhancing, wong2025adcd}, DCT traces improve robustness to recompression. However, such traces can also become a shortcut, causing the model to over-rely on DCT cues instead of RGB evidence. RiSE encodes the DCT coefficients $\mathbf{X}_{\mathrm{dct}}$ and quantization table $\mathbf{X}_{\mathrm{qt}}$ with the DCT encoder $\phi_{\mathrm{dct}}$ in \citep{qu2023towards}, and obtains DCT features $\mathbf{Z}_{\mathrm{dct}}=\phi_{\mathrm{dct}}(\mathbf{X}_{\mathrm{dct}},\mathbf{X}_{\mathrm{qt}})$. The DCT features then conditionally fuse with the RGB features using an image-level Bernoulli gate, $\mathbf{Z}_{\mathrm{fuse}}=\mathbf{Z}_{\mathrm{rgb}}+g\,\mathbf{Z}_{\mathrm{dct}}$ with $g\sim\mathrm{Bernoulli}(p_g)$, where $p_g=0.5$ is the default DCT fusion probability during training, and $p_g=1.0$ at inference. This random gate balances DCT and RGB learning by randomly dropping DCT traces during training, reducing reliance on DCT shortcuts. Unlike ADCD-Net \citep{wong2025adcd}, which uses a learnable DCT fusion score, our design does not assume DCT-grid alignment and avoids extra computation from alignment-breaking augmentations and score-learning modules. We observe that RiSE remains robust to such auxiliary cues despite freezing the principal components of an RGB-pretrained VFM, and future work could extend this fusion to noise prints \citep{guillaro2023trufor}, boundary traces \citep{liu2024attentive}, or edit traces \citep{wu2026editprint}.

\paragraph{Mask module.}
The mask module follows a Mask2Former-style design \citep{cheng2022masked}. With the VFM outputs $(\mathbf{Q}_\mathrm{vfm},\mathbf{Z}_\mathrm{vfm})$, the class head predicts per-query logits $\mathbf{C}=\phi_{\mathrm{cls}}(\mathbf{Q}_\mathrm{vfm})\in\mathbb{R}^{N_q\times 2}$. For mask prediction, we define the transformed query features $\mathbf{Q}_\mathrm{mask}=\phi_{\mathrm{mlp}}(\mathbf{Q}_\mathrm{vfm})\in\mathbb{R}^{N_q\times d}$ and the transformed dense features $\mathbf{Z}_\mathrm{mask}=\phi_{\mathrm{up}}(\mathbf{Z}_\mathrm{vfm})\in\mathbb{R}^{H_p \times W_p \times d}$. The mask head then predicts mask logits $\mathbf{M}\in\mathbb{R}^{N_q\times H_p\times W_p}$, where each $\mathbf{M}[q] \in \mathbb{R}^{H_p\times W_p}$ is obtained by the dot product between $\mathbf{Q}_\mathrm{mask}[q]$ and $\mathbf{Z}_\mathrm{mask}$. We then select $q^*=\arg\max_q (\mathbf{C}[q:0])$ and output $\hat{\mathbf{P}}=\sigma(\mathrm{Up}(\mathbf{M}[q^*])) \in [0,1]^{H\times W}$, where $\mathrm{Up}(\cdot)$ bilinearly upsamples the mask to $H\times W$, and $\sigma$ is the sigmoid function.
    
\paragraph{Training objective.}
We adopt the same Mask2Former \citep{cheng2022masked} losses: the class logits use Cross-Entropy (CE), and the mask logits use Binary Cross-Entropy (BCE) and Dice loss. The total loss is $ \mathcal{L}=\lambda_{\mathrm{CE}}\mathcal{L}_{\mathrm{CE}}+\lambda_{\mathrm{BCE}}\mathcal{L}_{\mathrm{BCE}}+\lambda_{\mathrm{Dice}}\mathcal{L}_{\mathrm{Dice}}$.

\subsection{RiSE Inference}

\begin{figure}
  \centering
  \includegraphics[width=0.8\linewidth]{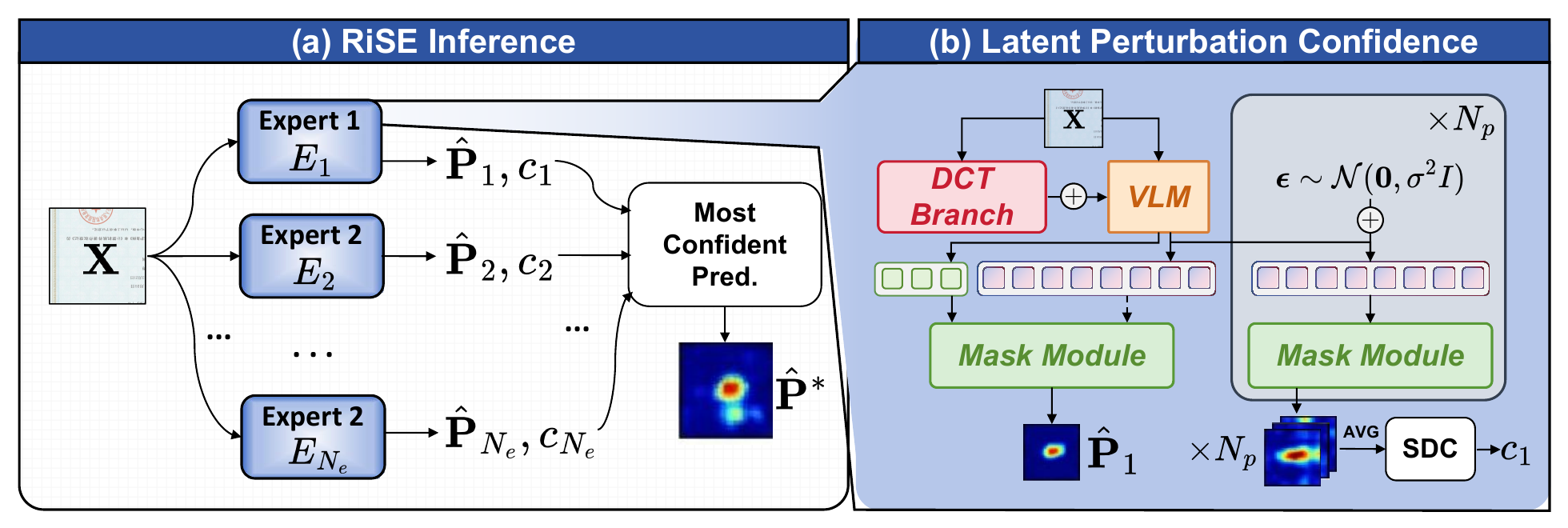}
    \caption{Inference overview of RiSE. (a) RiSE evaluates all experts in parallel on the same input and selects the final localization map from the most confident expert. (b) Latent Perturbation Confidence perturbs the latent features after the VFM backbone, averages the resulting mask predictions, and scores the averaged output with SDC for expert selection.}
    \label{fig:infer}
\end{figure}

As suggested by the expert-scaling trend in Fig.~\ref{fig:expert_scaling}~(b), adding residual experts improves cross-domain oracle F1. For a test set $\mathcal{D}_T=\{(\mathbf{X}_i,\mathbf{Y}_i)\}_{i=1}^{N_t}$ and evaluation metric $m$ (e.g. F1 score), we define the oracle score $\mathrm{Oracle}(\mathcal{D}_T, \{E_j\}_{j=1}^{N_e})=\frac{1}{N_t}\sum_{i=1}^{N_t}\max_{j\in\{1,\dots,N_e\}} m\bigl(E_j(\mathbf{X}_i),\mathbf{Y}_i\bigr)$ for RiSE with multiple experts. Because the ground-truth mask is unavailable at test time, this oracle score serves only as an upper bound on RiSE performance. As shown in Fig.~\ref{fig:infer}~(a), RiSE evaluates all experts in parallel on an input $\mathbf{X}$, producing expert-wise probability maps $\{\hat{\mathbf{P}}_j\}_{j=1}^{N_e}$. We then estimate a confidence score $c_j$ for each expert and select the most confident one, $j^*=\arg\max_j c_j$, as the final output. We evaluate three confidence measurements for this purpose, the existing Maximum Softmax Probability (MSP) \citep{hendrycks2017baseline} and Soft Dice Confidence (SDC) \citep{borges2024soft}, as well as our proposed Latent Perturbation Confidence (LPC). For clarity, we flatten the probability map $\hat{\mathbf{P}}$ into a vector $\hat{\mathbf{p}}\in[0,1]^{N_{\mathrm{pix}}}$, where $N_{\mathrm{pix}}=HW$.

\paragraph{MSP.}
As a simple baseline, we compute the average Maximum Softmax Probability (MSP) across all pixels \citep{hendrycks2017baseline}, $\mathrm{MSP}(\hat{\mathbf{p}})=\frac{1}{N_{\mathrm{pix}}}\sum_{n=1}^{N_{\mathrm{pix}}}\max\{\hat{\mathbf{p}}(n),\,1-\hat{\mathbf{p}}(n)\}$.
However, MSP can be dominated by the easy pristine pixels and underestimate the influence of tiny tampered regions. This motivates the usage of SDC, which focuses on the foreground confidence.

\paragraph{SDC.}
To address the extreme class imbalance, Soft Dice Confidence (SDC) \citep{borges2024soft} does not consider true negative predictions and therefore ignores the huge easy pristine area:
$\mathrm{SDC}(\hat{\mathbf{p}})= 2\hat{\mathbf{p}}^\top\hat{\mathbf{y}} / (\|\hat{\mathbf{p}}\|_1+\|\hat{\mathbf{y}}\|_1)$, where $\hat{\mathbf{y}}=\mathds{1}[\hat{\mathbf{p}}>\tau]\in\{0,1\}^{N_{\mathrm{pix}}}$, in which $\tau$ is the binarization threshold. SDC favors experts that consistently produce high probabilities for the forged area. Although SDC improves over MSP, it still lags behind oracle performance.

\paragraph{LPC.}
Our proposed LPC builds on SDC with a simple intuition: an expert should be trusted when its confidences remain stable under perturbed inputs. We therefore compute SDC on the prediction obtained by averaging over multiple Test-Time Augmentations (TTA), and Appendix~\ref{sec:tta_sdc_stability} provides a justification for why TTA averaging stabilizes SDC. Formally, let $T(\cdot)$ be an augmentation (e.g. flipping, rotation, noise), and the TTA-averaged prediction be $\bar{\mathbf{p}} = \frac{1}{M}\sum_{m=1}^{M}T_m^{-1}\!\left(E(T_m(\mathbf{X}))\right)$, where $T_m^{-1}$ maps the prediction back to the original coordinates, and for non-spatial perturbations such as additive Gaussian noise, $T_m^{-1}$ is the identity map. We then compute SDC on $\bar{\mathbf{p}}$ to obtain the TTA-SDC score. However, this input-space TTA is computationally expensive, yielding $M$ forward passes per image. To avoid the cost of TTA, we propose to apply perturbations directly to the latent space, that is the VFM output feature $\mathbf{Z}_\mathrm{vfm}$:
\[
\mathrm{LPC}(\mathbf{Z}_\mathrm{vfm})=\mathrm{SDC}\bigl(\bar{\mathbf{p}}_{\mathrm{lat}}\bigr),
\qquad
\bar{\mathbf{p}}_{\mathrm{lat}} = \frac{1}{M}\sum_{m=1}^{M} T_m^{-1}\!\left(\phi_{\mathrm{mask}}\bigl(\mathbf{Q}_\mathrm{vfm}, T_m(\mathbf{Z}_\mathrm{vfm})\bigr)\right),
\]
where $\phi_\mathrm{mask}$ is the mask module. Compared with image-space TTA, LPC avoids repeated VFM forwards by perturbing only the latent features. For Gaussian latent perturbations,
\[
T_m(\mathbf{Z}_\mathrm{vfm})=\mathbf{Z}_\mathrm{vfm}+\boldsymbol{\epsilon}_m, 
\qquad
\boldsymbol{\epsilon}_m\sim\mathcal{N}(\mathbf{0},\sigma^2 I). 
\]
The averaged output can then be written as the expected latent-perturbed output: 
\[
\bar{\mathbf{p}}_{\mathrm{lat}}=\mathbb{E}_{\boldsymbol{\epsilon}\sim\mathcal{N}(\mathbf{0},\sigma^2 I)}\bigl[\phi_{\mathrm{mask}}(\mathbf{Q}_\mathrm{vfm}, \mathbf{Z}_\mathrm{vfm}+\boldsymbol{\epsilon})\bigr].
\]
We find that this expectation admits an analytic approximation, without explicitly sampling the $M$ perturbations. Concretely, we propagate the mean and variance through the mask module using Assumed Density Filtering (ADF) \citep{gast2018lightweight} and details are provided in Appendix~\ref{app:adf_mask_module}. As a result, LPC requires only a single forward pass through both the VFM and the mask module.

\section{Experiments}

\paragraph{Dataset.}
We adapt the \emph{Doc Protocol} in ForensicHub \citep{du2025forensichub} for the main experiments. All models are trained only on the DocTamper training split \citep{qu2023towards} and are evaluated on seven test sets: three in-domain DocTamper test sets, \emph{FCD}, \emph{SCD}, and \emph{Test}, and four cross-domain test sets, \emph{T-SROIE} \citep{wang2022tampered}, \emph{OSTF} \citep{qu2025revisiting}, \emph{TPIC-13}\citep{wang2022detecting}, and \emph{RTM} \citep{luo2025toward}. Details of the datasets are provided in Appendix Table~\ref{tab:dataset_summary}.

\paragraph{Competitors.}
We compare RiSE with six general image forgery localization methods: MVSS-Net \citep{chen2021image}, PSCC-Net \citep{liu2022pscc}, CAT-Net \citep{kwon2022learning}, TruFor \citep{guillaro2023trufor}, IML-ViT \citep{ma2023iml}, and Mesorch \citep{zhu2025mesoscopic}, as well as five recent TFL methods: DTD \citep{qu2023towards}, FFDN \citep{chen2024enhancing}, CAFTB \citep{song2025cross}, TIFDM \citep{dong2024robust}, and ADCD-Net \citep{wong2025adcd}.

\paragraph{Implementation.}
RiSE uses DINOv3 ViT-L/16 \citep{simeoni2025dinov3} as the VFM backbone, with $L_1=10$, $L_2=10$, and $L_3=4$ blocks. The RiSE design choices are detailed in the ablation study in Table~\ref{tab:ablation_rise_design}. We train RiSE for 10,000 steps with the AdamW optimizer, using a learning rate of $3e-4$ decayed to $1e-5$ via cosine annealing, and a batch size of 16. Detailed training and inference efficiency are provided in Appendix Table~\ref{tab:efficiency}.

\subsection{Main Results}

\begin{table*}[t]
    \footnotesize
    \refstepcounter{table}
    {\footnotesize\raggedright\textbf{Table \thetable.} F1 scores on 7 TFL test sets. Number colors run from blue (lowest) to orange (highest) across this table; alternating rows have a light-gray background. Best is \textbf{bold} and second-best is \underline{underlined}. AvgI: average F1 on the 3 in-domain test sets. AvgC: average F1 on the 4 cross-domain test sets. Avg: average F1 on all 7 test sets. RiSE-$x$: $x$ experts without DCT traces. RiSE-9-DCT: 9 experts with DCT traces.\par}
    \label{tab:main_results}
    \centering
    \resizebox{\textwidth}{!}{%
    \rowcolors{3}{TableStripe}{white}
    \begin{tabular}{llcccccccccc}
        \toprule
        Methods & Venue & \multicolumn{4}{c}{In-Domain (Synthetic Data)} & \multicolumn{5}{c}{Cross-Domain (Real Data)} & Avg \\
        \cmidrule(lr){3-6} \cmidrule(lr){7-11}
        & & \emph{FCD} & \emph{SCD} & \emph{Test} & AvgI & \emph{T-SROIE} & \emph{OSTF} & \emph{TPIC-13} & \emph{RTM} & AvgC & \\
        \midrule
        TruFor & CVPR'23 & \scorecell{29}{0.261} & \scorecell{35}{0.312} & \scorecell{28}{0.252} & \scorecell{31}{0.275} & \scorecell{24}{0.213} & \scorecell{5}{0.046} & \scorecell{11}{0.104} & \scorecell{3}{0.034} & \scorecell{11}{0.099} & \scorecell{19}{0.175} \\
        MVSS-Net & ICCV'23 & \scorecell{23}{0.207} & \scorecell{42}{0.371} & \scorecell{43}{0.381} & \scorecell{36}{0.320} & \scorecell{21}{0.187} & \scorecell{4}{0.037} & \scorecell{12}{0.113} & \scorecell{2}{0.027} & \scorecell{10}{0.091} & \scorecell{21}{0.189} \\
        Mesorch & AAAI'25 & \scorecell{48}{0.423} & \scorecell{52}{0.459} & \scorecell{53}{0.465} & \scorecell{51}{0.449} & \scorecell{33}{0.294} & \scorecell{15}{0.139} & \scorecell{27}{0.241} & \scorecell{4}{0.041} & \scorecell{20}{0.178} & \scorecell{33}{0.294} \\
        IML-ViT & arXiv & \scorecell{53}{0.469} & \scorecell{58}{0.512} & \scorecell{51}{0.449} & \scorecell{54}{0.476} & \scorecell{48}{0.427} & \scorecell{23}{0.210} & \scorecell{29}{0.256} & \scorecell{8}{0.076} & \scorecell{27}{0.242} & \scorecell{39}{0.343} \\
        PSCC-Net & TCSVT'22 & \scorecell{44}{0.386} & \scorecell{44}{0.393} & \scorecell{56}{0.497} & \scorecell{48}{0.425} & \scorecell{59}{0.517} & \scorecell{50}{0.441} & \scorecell{62}{0.550} & \scorecell{14}{0.126} & \scorecell{46}{0.408} & \scorecell{47}{0.416} \\
        Cat-Net & IJCV'22 & \scorecell{87}{0.760} & \scorecell{73}{0.641} & \scorecell{87}{0.764} & \scorecell{82}{0.722} & \scorecell{69}{0.609} & \scorecell{20}{0.178} & \scorecell{39}{0.343} & \scorecell{7}{0.063} & \scorecell{34}{0.298} & \scorecell{54}{0.480} \\
        \midrule
        TIFDM & TCE'24 & \scorecell{10}{0.090} & \scorecell{29}{0.257} & \scorecell{29}{0.259} & \scorecell{23}{0.202} & \scorecell{6}{0.058} & \scorecell{0}{0.006} & \scorecell{1}{0.013} & \scorecell{1}{0.018} & \scorecell{2}{0.024} & \scorecell{11}{0.100} \\
        CAFTB & TOMM'24 & \scorecell{33}{0.292} & \scorecell{43}{0.377} & \scorecell{37}{0.328} & \scorecell{37}{0.332} & \scorecell{29}{0.262} & \scorecell{13}{0.119} & \scorecell{34}{0.301} & \scorecell{3}{0.033} & \scorecell{20}{0.179} & \scorecell{27}{0.244} \\
        DTD & CVPR'23 & \scorecell{78}{0.686} & \bestscore{84}{0.739} & \secondscore{92}{0.803} & \scorecell{85}{0.743} & \scorecell{60}{0.525} & \scorecell{14}{0.124} & \scorecell{32}{0.284} & \scorecell{6}{0.058} & \scorecell{28}{0.247} & \scorecell{52}{0.460} \\
        FFDN & ECCV'24 & \bestscore{100}{0.877} & \bestscore{84}{0.739} & \bestscore{94}{0.821} & \bestscore{93}{0.813} & \scorecell{61}{0.533} & \scorecell{27}{0.241} & \scorecell{40}{0.357} & \scorecell{7}{0.071} & \scorecell{34}{0.300} & \scorecell{59}{0.520} \\
        ADCD-Net & ICCV'25 & \scorecell{92}{0.803} & \secondscore{82}{0.718} & \scorecell{90}{0.791} & \secondscore{88}{0.771} & \scorecell{71}{0.623} & \scorecell{50}{0.441} & \scorecell{66}{0.579} & \scorecell{18}{0.159} & \scorecell{51}{0.450} & \scorecell{67}{0.588} \\
        \midrule
        RiSE-1 & Ours & \scorecell{85}{0.744} & \scorecell{67}{0.590} & \scorecell{74}{0.654} & \scorecell{75}{0.662} & \scorecell{83}{0.728} & \scorecell{68}{0.597} & \scorecell{78}{0.686} & \scorecell{17}{0.158} & \scorecell{62}{0.542} & \scorecell{68}{0.594} \\
        RiSE-3 & Ours & \scorecell{87}{0.764} & \scorecell{67}{0.588} & \scorecell{77}{0.675} & \scorecell{77}{0.676} & \scorecell{87}{0.760} & \scorecell{71}{0.625} & \scorecell{84}{0.740} & \scorecell{19}{0.173} & \scorecell{65}{0.575} & \scorecell{70}{0.618} \\
        RiSE-9 & Ours & \scorecell{90}{0.791} & \scorecell{66}{0.579} & \scorecell{79}{0.692} & \scorecell{78}{0.687} & \secondscore{89}{0.785} & \bestscore{79}{0.692} & \bestscore{91}{0.795} & \secondscore{19}{0.175} & \secondscore{69}{0.611} & \secondscore{73}{0.644} \\
        RiSE-9-DCT & Ours & \secondscore{99}{0.864} & \scorecell{75}{0.662} & \scorecell{88}{0.775} & \scorecell{87}{0.767} & \bestscore{92}{0.808} & \secondscore{74}{0.654} & \secondscore{89}{0.784} & \bestscore{24}{0.218} & \bestscore{70}{0.616} & \bestscore{77}{0.681} \\
        \bottomrule
    \end{tabular}%
    }
\end{table*}

Table~\ref{tab:main_results} highlights the superiority of RiSE over prior TFL methods, especially under domain shift. Compared with ADCD-Net, RiSE-1 improves cross-domain F1 by 20.4\%, RiSE-3 increases this gain to 27.8\%, and RiSE-9 further enlarges the cross-domain F1 to 35.8\%. Prior non-DCT detectors fall short in both domains. PSCC-Net reaches a relatively strong cross-domain F1 of 0.408, yet the in-domain F1 remains only 0.425. Adding DCT cues improves in-domain fitting for Cat-Net, DTD, FFDN, and ADCD-Net, all with in-domain F1 above 0.7, but the best cross-domain F1 still peaks at only 0.450 (ADCD-Net). RiSE-9-DCT obtains an in-domain F1 of 0.767, a cross-domain F1 of 0.616, and an overall F1 of 0.681. Relative to ADCD-Net, RiSE-9-DCT offers 36.8\% higher cross-domain F1 and 15.7\% higher overall F1. Compared with RiSE-9, RiSE-9-DCT increases AvgI from 0.687 to 0.767 and increases AvgC from 0.611 to 0.616. RTM is particularly challenging because it contains manual manipulations, and the oracle F1 of RiSE-9 in Fig.~\ref{fig:expert_scaling}~(b) can further raise the RTM F1 to 0.277, encouraging more advanced expert-selection methods in future studies.

\subsection{Ablation Study}

\noindent
\begin{minipage}[h]{0.49\linewidth}
    \vspace{0pt}
    \refstepcounter{table}
    {\footnotesize\textbf{Table \thetable.} Ablation on RiSE-9 design. $\ast$: LoRA. FT: fine-tune blocks. A: attention. M: MLP. Shared: shared mask module. $N_q$: number of queries. $p_g$: DCT fusion probability. Router: expert routing. $\dagger$: two-stage MoE. \textbf{RiSE-9 uses \#16 setting. RiSE-9-DCT uses \#18 setting.} \par}
    \label{tab:ablation_rise_design}
    \centering
    \tiny
    \setlength{\tabcolsep}{2pt}
    \resizebox{\linewidth}{!}{%
    \rowcolors{2}{TableStripe}{white}
    \begin{tabular}{l c c c c c c c c c}
        \toprule
        \# & Rank & FT & Shared & $N_q$ & $p_g$ & Router & AvgI & AvgC & Avg \\
        \midrule
        \#1 & \multicolumn{6}{c}{Full-parameter Fine-tune} & \scorecell{69}{0.541} & \scorecell{0}{0.211} & \scorecell{30}{0.353} \\
        \midrule
        \#2 & 1 & A + M & \checkmark & 10 & 0 & Offline & \scorecell{42}{0.411} & \scorecell{44}{0.423} & \scorecell{43}{0.418} \\
        \#3 & 1 & A + M & \checkmark & 20 & 0 & Offline & \scorecell{68}{0.539} & \scorecell{43}{0.416} & \scorecell{54}{0.469} \\
        \midrule
        \#4 & 1 & A + M &  & 1 & 0 & Offline & \scorecell{89}{0.637} & \scorecell{68}{0.535} & \scorecell{77}{0.579} \\
        \midrule
        \#5 & 1 & A + M & \checkmark & 1 & 0 & $\text{MoE}^{\dagger}$ & \scorecell{98}{0.682} & \scorecell{47}{0.435} & \scorecell{69}{0.541} \\
        \#6 & 1 & A + M & \checkmark & 1 & 0 & MoE & \scorecell{99}{0.686} & \scorecell{69}{0.540} & \scorecell{82}{0.603} \\
        \midrule
        \#7 & 1 & A & \checkmark & 1 & 0 & Offline & \scorecell{84}{0.613} & \scorecell{76}{0.573} & \scorecell{79}{0.590} \\
        \#8 & 1 & M & \checkmark & 1 & 0 & Offline & \scorecell{92}{0.654} & \scorecell{77}{0.580} & \scorecell{84}{0.612} \\
        \midrule
        \#9 & $\text{16}^\ast$ & A + M & \checkmark & 1 & 0 & Offline & \scorecell{93}{0.658} & \scorecell{80}{0.596} & \scorecell{86}{0.623} \\
        \#10 & $\text{8}^\ast$ & A + M & \checkmark & 1 & 0 & Offline & \scorecell{94}{0.663} & \scorecell{78}{0.587} & \scorecell{85}{0.620} \\
        \#11 & $\text{4}^\ast$ & A + M & \checkmark & 1 & 0 & Offline & \scorecell{94}{0.659} & \scorecell{79}{0.589} & \scorecell{85}{0.619} \\
        \#12 & $\text{1}^\ast$ & A + M & \checkmark & 1 & 0 & Offline & \scorecell{93}{0.656} & \scorecell{78}{0.587} & \scorecell{85}{0.617} \\
        \midrule
        \#13 & 16 & A + M & \checkmark & 1 & 0 & Offline & \scorecell{95}{0.666} & \scorecell{78}{0.587} & \scorecell{86}{0.621} \\
        \#14 & 8 & A + M & \checkmark & 1 & 0 & Offline & \scorecell{95}{0.665} & \scorecell{77}{0.580} & \scorecell{85}{0.616} \\
        \#15 & 4 & A + M & \checkmark & 1 & 0 & Offline & \scorecell{95}{0.665} & \scorecell{80}{0.593} & \scorecell{86}{0.624} \\
                \#16 & 1 & A + M & \checkmark & 1 & 0 & Offline & \scorecell{94}{0.663} & \scorecell{80}{0.596} & \scorecell{86}{0.625} \\
        \midrule
        \#17 & 1 & A + M & \checkmark & 1 & 1 & Offline & \scorecell{91}{0.648} & \scorecell{64}{0.519} & \scorecell{76}{0.574} \\
                \#18 & 1 & A + M & \checkmark & 1 & 0.5 & Offline & \scorecell{100}{0.690} & \scorecell{77}{0.581} & \scorecell{87}{0.628} \\
        \bottomrule
    \end{tabular}%
    }
\end{minipage}\hfill
\begin{minipage}[h]{0.48\linewidth}
    \vspace{0pt}
    \paragraph{RiSE design.}
    Table~\ref{tab:ablation_rise_design} ablates the main RiSE-9 design choices using 5,000 training steps. Among the non-DCT settings ($p_g=0$), \#16 performs the best, achieving an Avg. F1 of 0.625, and we use \#16 as the default setting of RiSE-9. Compared with full-parameter fine-tuning (\#1), \#16 improves Avg. F1 by 77.1\%, showing the superior performance of residual subspace adaptation. Increasing the number of queries to 10 or 20 (\#2--\#3) reduces Avg. F1 by 33.1\% and 25.0\%, respectively. Making the mask module expert-specific (\#4) causes a 7.4\% drop, indicating that stronger parameter sharing is beneficial. Replacing offline assignment with standard MoE (\#6) or two-stage MoE (\#5), where experts are trained first and the router is trained afterward, lowers Avg. F1 by 3.5\% and 13.4\%, respectively. Using only attention adaptation (\#7) or only MLP adaptation (\#8) reduces Avg. F1 by 5.6\% and 2.1\%, suggesting that both adaptations help, with MLP being slightly more important. The residual-rank study further supports a lightweight design. Increasing the residual rank to 4, 8, and 16 (\#15, \#14, \#13) reduces Avg. F1 by 0.2\%, 1.4\%, and
\end{minipage}

0.6\%, while the LoRA-style variants (\#9--\#12) are also consistently worse, with drops of 0.3\%, 0.8\%, 1.0\%, and 1.3\%. After introducing DCT traces, always enabling fusion during training ($p_g=1$, \#17) reduces AvgI from 0.663 to 0.648 and AvgC from 0.596 to 0.519, a 12.9\% drop relative to \#16. In contrast, random-gated fusion with $p_g=0.5$ (\#18) boost the AvgI to 0.690 and the AvgC to 0.581, with a Avg. F1 of 0.628 which is 0.5\% higher than \#16.

\begin{minipage}[t]{0.52\linewidth}
    \vspace{0pt}
    \footnotesize
    \refstepcounter{table}
    {\footnotesize\textbf{Table \thetable.} Ablation on confidence estimation for RiSE-9. F\&R: Flip-and-Rotation TTA; GP: Latent Gaussian Perturbation; AGP: Analytic Gaussian Propagation. \par}
    \label{tab:ablation_confidence}
    \centering
    \setlength{\tabcolsep}{2pt}
    \rowcolors{2}{TableStripe}{white}
    \begin{tabular}{lccccc}
        \toprule
        Method & \emph{T-SROIE} & \emph{OSTF} & \emph{TPIC-13} & \emph{RTM} & AvgC \\
        \midrule
        Mean Expert & \scorecell{78}{0.688} & \scorecell{25}{0.282} & \scorecell{28}{0.311} & \scorecell{0}{0.094} & \scorecell{33}{0.344} \\
        MSP & \scorecell{91}{0.789} & \scorecell{59}{0.548} & \scorecell{79}{0.698} & \scorecell{0}{0.093} & \scorecell{57}{0.532} \\
        SDC & \secondscore{91}{0.790} & \scorecell{60}{0.551} & \scorecell{79}{0.698} & \scorecell{0}{0.096} & \scorecell{58}{0.534} \\
        TTA-SDC (F\&R) & \scorecell{91}{0.789} & \scorecell{77}{0.679} & \scorecell{90}{0.785} & \scorecell{9}{0.159} & \scorecell{67}{0.603} \\
        LPC (F\&R) & \scorecell{86}{0.748} & \scorecell{76}{0.674} & \scorecell{90}{0.780} & \scorecell{8}{0.158} & \scorecell{65}{0.590} \\
        LPC (GP) & \scorecell{90}{0.784} & \scorecell{78}{0.689} & \secondscore{92}{0.796} & \secondscore{11}{0.175} & \secondscore{68}{0.611} \\
        LPC (AGP) & \scorecell{90}{0.785} & \secondscore{78}{0.692} & \scorecell{92}{0.795} & \secondscore{11}{0.175} & \secondscore{68}{0.611} \\
        Oracle & \bestscore{100}{0.859} & \bestscore{87}{0.757} & \bestscore{98}{0.843} & \bestscore{24}{0.277} & \bestscore{77}{0.684} \\
        \bottomrule
    \end{tabular}%
\end{minipage}\hfill
\begin{minipage}[t]{0.46\linewidth}
    \vspace{0pt}
    \centering
    \includegraphics[width=\linewidth]{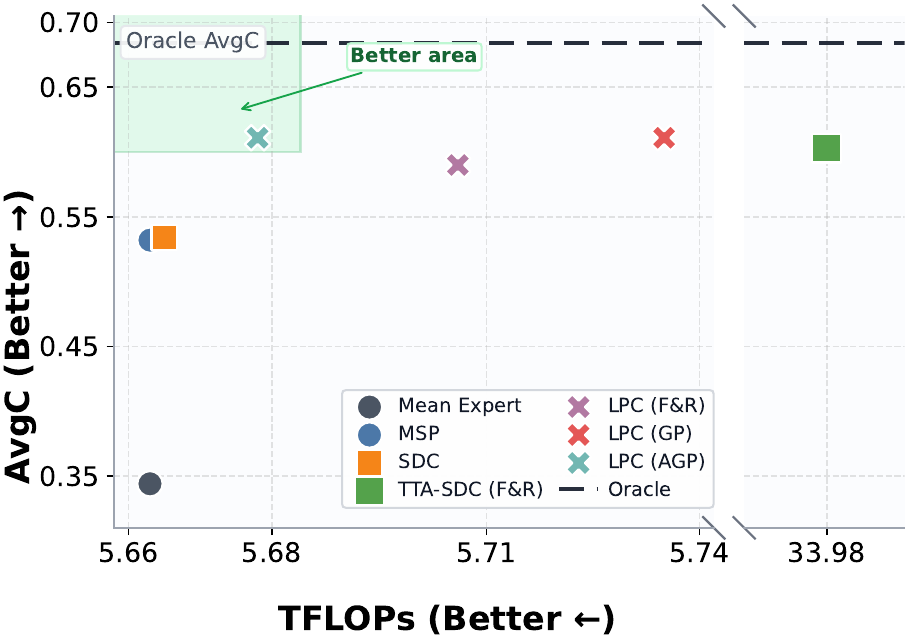}
    \refstepcounter{figure}
    {\footnotesize\textbf{Figure \thefigure.} Cross-domain F1 versus inference TFLOPs.}
    \label{fig:avgc_tflops_scatter}
\end{minipage}

\begin{minipage}[h]{0.52\linewidth}
    \vspace{0pt}
    \paragraph{Expert confidence estimation.}
    Table~\ref{tab:ablation_confidence} compares expert-selection strategies for RiSE-9 on the 4 cross-domain test sets, while Fig.~\ref{fig:avgc_tflops_scatter} summarizes the corresponding Avg. F1 and TFLOPs comparison. A simple mean over expert predictions performs poorly with 0.344 F1, while MSP and SDC improve F1 to 0.532 and 0.534, respectively. Input-space TTA-SDC with five Flip-and-Rotation (F\&R) augmentations (horizontal flip, vertical flip, and clockwise rotations by $90^\circ$, $180^\circ$, and $270^\circ$) boosts Avg. F1 to 0.603. This supports the view that TTA stabilizes confidence estimation, but it requires $6\times$ total computation because the model is evaluated on the original input plus 5 augmented samples, as shown in Fig.~\ref{fig:avgc_tflops_scatter}. Our proposed LPC (F\&R) achieves a similar Avg. F1 of 
\end{minipage}\hfill
\begin{minipage}[h]{0.46\linewidth}
    \vspace{0pt}  
    \centering
    \includegraphics[width=\linewidth]{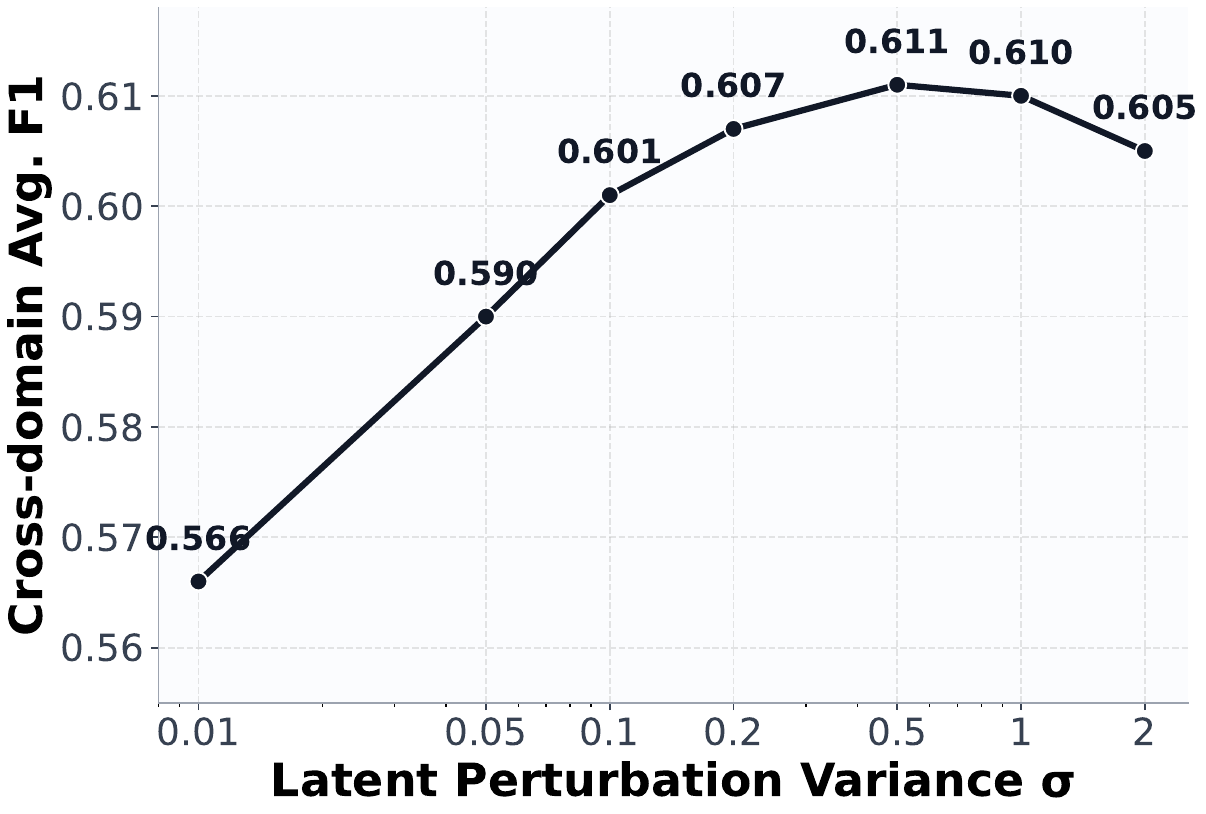}
    \refstepcounter{figure}
    {\footnotesize\textbf{Figure \thefigure.} Effect of the latent perturbation variance on cross-domain performance.}
    \label{fig:lpc_variance_line}
\end{minipage}

0.590 with the same F\&R augmentations, but it is more efficient because it applies perturbations in the latent space and does not require repeated VFM forwards. LPC (GP) with 10 latent Gaussian perturbations reaches the same Avg. F1 of 0.611 as our analytic Gaussian propagation LPC (AGP), but requires more computation. LPC (AGP) is especially attractive because it preserves single-pass inference and closes more of the gap to the oracle selector without incurring the compute growth of TTA. Fig.~\ref{fig:lpc_variance_line} further examines the effect of the latent perturbation variance ($\sigma^2$) on cross-domain performance. The trend shows that a moderate perturbation variance ($\sigma^2=0.5$) gives the best confidence estimation, whereas too little noise provides limited stability information and too much noise degrades the localization signal. 

\section{Conclusion}

We presented RiSE, a residual subspace expert model for text-centric image forgery localization. Motivated by the observation that FPFT causes low-rank feature collapse and poor cross-domain generalization, RiSE freezes the dominant VFM subspace, adapts only a small residual component, and scales capacity through independently trained forgery-aware experts. We also introduced LPC, a confidence estimator that selects experts by measuring prediction stability under latent perturbations via a single-pass approximation. Experiments show that RiSE consistently improves cross-domain localization as the number of experts increases. In particular, relative to ADCD-Net, RiSE-1 improves AvgC by 20.4\%, and scaling to RiSE-9 increases the gain to 35.8\%. RiSE-9-DCT obtains an in-domain average F1 of $0.767$, a cross-domain average F1 of $0.616$, and an overall average F1 of $0.681$. These results suggest that preserving transferable VFM priors while scaling expertized residual capacity is a promising direction for robust and efficient TFL.

\section*{Acknowledgements}

This work was supported in part by Macau Science and Technology Development Fund under 001/2024/SKL, 0119/2024/RIB2 and 0110/2025/R1B2; in part by Research Committee at University of Macau under MYRG-CRG2025-00031-FST and MYRG-GRG2025-00086-FST; in part by the Guangdong Basic and Applied Basic Research Foundation under Grant 2024A1515012536.

\bibliographystyle{plainnat}
\bibliography{ref}

\clearpage


\appendix
\appendix

\section{Dataset Summary}

\begin{table*}[ht]
    \caption{Summary of datasets used in our experiments.}
    \centering
    \setlength{\tabcolsep}{4pt}
    \resizebox{\textwidth}{!}{%
    \rowcolors{2}{TableStripe}{white}
    \begin{tabular}{p{4.5cm}p{1.3cm}p{2.3cm}p{1.5cm}p{6.7cm}}
        \toprule
        Dataset & Usage & Setting & \#Sample & Description \\
        \midrule
        \emph{Training Set} \citep{qu2023towards} & Train & Source domain & 120,000 & A synthetic training set of images with copy-move, splicing, and print-based edits on contracts, receipts, and books. \\
        \emph{Test} \citep{qu2023towards} & Test & In-domain & 30,000 & A general synthetic test set with copy-move, splicing, and print-based edits. \\
        \emph{FCD} \citep{qu2023towards} & Test & In-domain & 2,000 & A synthetic test set with copy-move, splicing, and print-based edits, whose source images come from the Noisy Office Dataset. \\
        \emph{SCD} \citep{qu2023towards} & Test & In-domain & 18,000 & A synthetic test set with copy-move, splicing, and print-based edits, whose source images come from Huawei Cloud. \\
        \emph{T-SROIE} \citep{wang2022tampered} & Test & Cross-domain & 1,579 & Scanned receipts tampered using the AIGC text editing model SR-Net. \\
        \emph{OSTF} \citep{qu2025revisiting} & Test & Cross-domain & 3,046 & Natural scene text images tampered using eight different AIGC-based text editing models. \\
        \emph{TPIC-13} \citep{wang2022detecting} & Test & Cross-domain & 589 & Naturally captured scene-text images tampered using the AIGC text editing model SR-Net. \\
        \emph{RTM} \citep{luo2025toward} & Test & Cross-domain & 3,444 & Document images with both synthetic and manual manipulations, covering copy-move, splicing, print, and erasure edits across diverse document types such as scanned forms. \\
        \bottomrule
    \end{tabular}
    }
    \label{tab:dataset_summary}
\end{table*}

Following the \emph{Doc Protocol} in ForensicHub \citep{du2025forensichub}, we use the fully synthetic training set from DocTamper \citep{qu2023towards} to train all models and evaluate on seven test sets covering both in-domain and cross-domain scenarios. The summary of these datasets is shown in Table~\ref{tab:dataset_summary}. All samples in the in-domain \emph{Test}, \emph{FCD}, and \emph{SCD} splits are also synthetic DocTamper samples, compressed once using the final JPEG quality-factor records provided by DocTamper, while all samples in the cross-domain test sets are cropped to $512\times 512$ patches without additional compression. The training set and these three in-domain test sets share the same three forgery synthesis types: copy-move, splicing, and print-based edits. The cross-domain test sets have more diverse forgery sources, including AIGC-based text editing models and manual manipulation.

\section{Expert Assignment}
\label{app:domain_label_acquisition}
RiSE uses different training-set partitions for different numbers of experts. RiSE-1 uses the entire training set as a single subset. RiSE-3 uses the forgery synthesis type as the subset label, which splits the training set into three subsets corresponding to copy-move (CM), splicing (SP), and generation (GE). RiSE-9 uses a finer expert-assignment label $d_i$ for each training sample by combining two attributes, the forgery synthesis type $s_i$ and the feature cluster index $k_i$. The forgery synthesis type is provided directly by the DocTamper metadata (see \url{https://github.com/qcf-568/DocTamper}). Each training sample belongs to one of the three synthesis categories CM, SP, or GE. The training set contains 35{,}000 CM samples, 45{,}000 SP samples, and 40{,}000 GE samples. We use this label as the first-level partition because these three manipulation sources exhibit different forensic traces and therefore benefit from separate expert specialization.

The feature cluster index is obtained from the single output query embedding of a trained RiSE-1 model. Specifically, RiSE sets $N_q=1$, so for each training image the model produces one output query embedding, which we denote by $\mathbf{q}_i\in\mathbb{R}^{d}$. We use this single query embedding rather than spatially pooled dense features because pooled dense tokens can mix authentic and tampered regions, especially when the forged area is small, which entangles real and fake evidence in a single representation. In contrast, the output query embedding is optimized by the localization objective to summarize tampering-relevant evidence at the image level, making it a cleaner feature for clustering.

For each forgery type $s$, let $\mathcal{Q}_s=\{\mathbf{q}_i\mid s_i=s\}$. We first run standard K-means with $N_c$ clusters on $\mathcal{Q}_s$ to obtain initial cluster centers $\{\boldsymbol{\mu}_{s,k}\}_{k=1}^{N_c}$. To prevent a few clusters from dominating the training set, we then rebalance the assignments with a balanced-clustering step that approximately solves
\[
\min_{\{k_i\}}\;\sum_{i:s_i=s}\lVert \mathbf{q}_i-\boldsymbol{\mu}_{s,k_i}\rVert_2^2
\quad
\text{s.t.}\quad
\bigl|\{i:s_i=s,\,k_i=k\}\bigr|\in\left\{\left\lfloor\frac{|\mathcal{Q}_s|}{N_c}\right\rfloor,\left\lceil\frac{|\mathcal{Q}_s|}{N_c}\right\rceil\right\},\ \forall k.
\]
In practice, we implement this balancing with a greedy reassignment procedure. Samples are sorted by their nearest-center distance, and each sample is assigned to the closest cluster whose capacity has not yet been filled. If all preferred clusters are already full, the sample is assigned to the currently smallest cluster. This produces nearly equal cluster sizes within each forgery type while staying close to the original K-means solution.

After obtaining both attributes, we define the final expert-assignment label as
\[
d_i=(s_i-1)N_c+k_i,
\]
which yields $N_e=N_sN_c$ non-overlapping expert assignments.

\section{Effective Rank Computation}
\label{app:effective_rank}

We compute effective rank from the empirical PCA spectrum of the saved penultimate features of DTD, FFDN, ADCD-Net and our RiSE-1 with 100 samples per dataset. For each image, we spatially average the penultimate feature map into a single feature vector, then stack the resulting vectors into a feature matrix $\mathbf{F} \in \mathbb{R}^{100 \times d}$, where $d$ is the feature dimension. After centering $\mathbf{F}$, we compute singular values $\{\alpha_a\}$ and convert them into normalized spectral weights
\[
\rho_a = \frac{\eta_a}{\sum_b \eta_b}, \qquad \eta_a = \frac{\alpha_a^2}{100-1}.
\]
Following the entropy-based definition used in low-rank analysis, the effective rank is
\[
\mathrm{erank}(\mathbf{F})=\exp\left(-\sum_a \rho_a \log \rho_a\right).
\]
Intuitively, the effective rank measures how many principal directions are effectively used by the representation. If almost all variance is concentrated in one or two directions, the effective rank is close to $1$, such as DTD, FFDN, ADCD-Net in Fig.~\ref{fig:effective_rank_compare}. If variance is distributed more evenly across many directions, the effective rank increases, such as RiSE-1 in Fig.~\ref{fig:effective_rank_compare}.

\section{Why TTA Averaging Stabilizes the Soft Dice Confidence}
\label{sec:tta_sdc_stability}

In this section, we explain why averaging predictions over Test-Time
Augmentation (TTA) makes the Soft Dice Confidence (SDC) more stable than
using a single prediction. Let $\mathbf{X}$ be the input image, $E$ the expert
model, $T_m(\cdot)$ the $m$-th augmentation, $M$ the number of
augmentations, and $N_{\mathrm{pix}} = HW$ the number of pixels. For spatial
augmentations such as flips and rotations, the prediction must be mapped back
to the original image coordinates before averaging, which we denote by
$T_m^{-1}(\cdot)$. For non-spatial augmentations such as additive noise,
$T_m^{-1}$ is simply the identity map. The original prediction is the special
case corresponding to the identity augmentation. Let
\[
T_1,\dots,T_M\stackrel{\mathrm{i.i.d.}}{\sim}\mathcal{A}
\]
be label-preserving augmentations drawn from an augmentation distribution
$\mathcal{A}$. We denote the prediction for the $m$-th augmented output by $\hat{\mathbf{p}}_m$, the TTA average by $\bar{\mathbf{p}}_{M}$, and the ideal
augmentation expectation by $\bar{\mathbf{p}}_{\mathcal{A}}$:
\[
\hat{\mathbf{p}}_m:=T_m^{-1}\!\left(E\!\bigl(T_m(\mathbf{X})\bigr)\right)\in[0,1]^{N_{\mathrm{pix}}},
\quad
\bar{\mathbf{p}}_{M}:=\frac{1}{M}\sum_{m=1}^{M}\hat{\mathbf{p}}_m,
\quad
\bar{\mathbf{p}}_{\mathcal{A}}:=\mathbb{E}_{T\sim\mathcal{A}}\!\left[T^{-1}(E(T(\mathbf{X})))\right].
\]
    
\paragraph{Variance reduction.}
For each pixel index $n\in\{1,\dots,N_{\mathrm{pix}}\}$, let
$\sigma^{2}(n)=\mathrm{Var}_{T\sim\mathcal{A}}\!\bigl[\hat{\mathbf{p}}_m(n)\bigr]$.
Since $\bar{\mathbf{p}}_{\mathcal{A}}(n)=\mathbb{E}[\hat{\mathbf{p}}_m(n)]$ and
the $\hat{\mathbf{p}}_m$ are i.i.d., the variance of the sample mean at pixel
$n$ is
\[
\mathbb{E}\bigl[(\bar{\mathbf{p}}_{M}(n)-\bar{\mathbf{p}}_{\mathcal{A}}(n))^2\bigr]
=\mathrm{Var}\!\left(\frac{1}{M}\sum_{m=1}^{M}\hat{\mathbf{p}}_{m}(n)\right)
=\frac{1}{M^2}\sum_{m=1}^{M}\mathrm{Var}\!\bigl[\hat{\mathbf{p}}_{m}(n)\bigr]
=\frac{\sigma^{2}(n)}{M}.
\]
To extend this pointwise result to the full prediction vector, let
$\boldsymbol{\Delta}=\bar{\mathbf{p}}_{M}-\bar{\mathbf{p}}_{\mathcal{A}}$.
Using $\|\boldsymbol{\Delta}\|_1\le\sqrt{N_{\mathrm{pix}}}\,\|\boldsymbol{\Delta}\|_2$
and Jensen's inequality, we obtain
\[
\mathbb{E}\|\boldsymbol{\Delta}\|_1
\le \sqrt{N_{\mathrm{pix}}}\,\mathbb{E}\|\boldsymbol{\Delta}\|_2
\le \sqrt{N_{\mathrm{pix}}\,\mathbb{E}\|\boldsymbol{\Delta}\|_2^2}
=\sqrt{N_{\mathrm{pix}}\sum_{n=1}^{N_{\mathrm{pix}}}\mathbb{E}[\Delta(n)^2]}.
\]
Substituting the pixel-wise variance identity yields
\begin{equation}
\mathbb{E}\bigl[(\bar{\mathbf{p}}_{M}(n)-\bar{\mathbf{p}}_{\mathcal{A}}(n))^2\bigr]
=\frac{\sigma^{2}(n)}{M},
\qquad
\mathbb{E}\bigl\|\bar{\mathbf{p}}_{M}-\bar{\mathbf{p}}_{\mathcal{A}}\bigr\|_1
\le\sqrt{\frac{N_{\mathrm{pix}}\sum_{n=1}^{N_{\mathrm{pix}}}\sigma^{2}(n)}{M}}
=\mathcal{O}\!\bigl(M^{-1/2}\bigr),
\label{eq:p_var}
\end{equation}
i.e., the TTA average concentrates around its augmentation expectation at the rate of $M^{-1/2}$, whereas a single prediction $\hat{\mathbf{p}}_{m}$ retains the full variance $\sigma^{2}(n)$.

\paragraph{Lipschitz continuity of SDC.}
Define
\[
\mathrm{SDC}(\hat{\mathbf{p}},\hat{\mathbf{y}})=\frac{2\hat{\mathbf{p}}^\top\hat{\mathbf{y}}}{\|\hat{\mathbf{p}}\|_1+\|\hat{\mathbf{y}}\|_1},
\qquad
\hat{\mathbf{y}}=\mathds{1}[\hat{\mathbf{p}}>\tau].
\]
Conditioned on the hard predicted mask
$\hat{\mathbf{y}}\in\{0,1\}^{N_{\mathrm{pix}}}$, we can view
$\mathrm{SDC}(\hat{\mathbf{p}})$ as a function of $\hat{\mathbf{p}}$ alone.
Applying the quotient rule to the $n$-th entry $\hat{\mathbf{p}}(n)$ gives
\[
\frac{\partial\,\mathrm{SDC}(\hat{\mathbf{p}})}{\partial\hat{\mathbf{p}}(n)}
=\frac{2\hat{y}(n)\bigl(\|\hat{\mathbf{p}}\|_1+\|\hat{\mathbf{y}}\|_1\bigr)-2\hat{\mathbf{p}}^\top\hat{\mathbf{y}}}
    {(\|\hat{\mathbf{p}}\|_1+\|\hat{\mathbf{y}}\|_1)^{2}},
\qquad
\left|\frac{\partial\,\mathrm{SDC}(\hat{\mathbf{p}})}{\partial\hat{\mathbf{p}}(n)}\right|\le \frac{2}{\|\hat{\mathbf{p}}\|_1+\|\hat{\mathbf{y}}\|_1}.
\]
The inequality follows by considering the two cases $\hat{y}(n)=0$ and
$\hat{y}(n)=1$. In both cases, the numerator has magnitude at most
$2(\|\hat{\mathbf{p}}\|_1+\|\hat{\mathbf{y}}\|_1)$ because
$0\le \hat{\mathbf{p}}^\top\hat{\mathbf{y}}\le
\|\hat{\mathbf{p}}\|_1+\|\hat{\mathbf{y}}\|_1$. Therefore,
$\mathrm{SDC}(\cdot)$ is locally $L$-Lipschitz in $\ell_1$ along the segment
between $\bar{\mathbf{p}}_{M}$ and $\bar{\mathbf{p}}_{\mathcal{A}}$, with
$L\le 2/(\|\bar{\mathbf{p}}_{\mathcal{A}}\|_1+\|\hat{\mathbf{y}}\|_1)$ up to local
variation in the denominator. Hence,
\begin{equation}
\bigl|\mathrm{SDC}(\bar{\mathbf{p}}_{M})-\mathrm{SDC}(\bar{\mathbf{p}}_{\mathcal{A}})\bigr|
\;\le\; L\,\|\bar{\mathbf{p}}_{M}-\bar{\mathbf{p}}_{\mathcal{A}}\|_1.
\label{eq:sdc_lip}
\end{equation}

\paragraph{Variance of SDC.}
Combining \eqref{eq:p_var} and \eqref{eq:sdc_lip} yields
\begin{equation}
\mathbb{E}\bigl|\mathrm{SDC}(\bar{\mathbf{p}}_{M})-\mathrm{SDC}(\bar{\mathbf{p}}_{\mathcal{A}})\bigr|
\;\le\;
\frac{2}{\|\bar{\mathbf{p}}_{\mathcal{A}}\|_1+\|\hat{\mathbf{y}}\|_1}
\sqrt{\frac{N_{\mathrm{pix}}\sum_{n=1}^{N_{\mathrm{pix}}}\sigma^{2}(n)}{M}}
\;=\;\mathcal{O}\bigl(M^{-1/2}\bigr),
\label{eq:sdc_rate}
\end{equation}

Equation~\eqref{eq:sdc_rate} formalizes the stabilizing effect of TTA
averaging on SDC. The single-pass quantity $\mathrm{SDC}(\hat{\mathbf{p}})$
inherits the full augmentation-induced variability and can fluctuate
substantially from one augmentation to another. In contrast,
$\mathrm{SDC}(\bar{\mathbf{p}}_{M})$ is computed from an averaged prediction
whose deviation from the augmentation expectation decreases at the rate of $M^{-1/2}$. As a result, the expert ranking induced by
$\mathrm{SDC}(\bar{\mathbf{p}}_{M})$ is more stable to perturbations
than the ranking induced by $\mathrm{SDC}(\hat{\mathbf{p}})$.

\section{Moment Propagation in the Mask Module}
\label{app:adf_mask_module}

\renewcommand{\lstlistingname}{Pseudocode}
\renewcommand{\lstlistlistingname}{List of Pseudocodes}

This section gives the implementation details of the proposed Analytic Gaussian Propagation (AGP) used by the Latent Perturbation Confidence (LPC). We first summarize the deterministic forward pass of the mask module with the PyTorch-style Pseudocode~\ref{lst:mask_module_forward}, then give the Assumed Density Filtering (ADF) rules for propagating the mean and variance under latent Gaussian perturbations.

\paragraph{Mask module forward.}
For an image, the VFM produces output queries $\mathbf{Q}_\mathrm{vfm}\in\mathbb{R}^{N_q\times d}$ and dense tokens $\mathbf{Z}_\mathrm{vfm}\in\mathbb{R}^{H_pW_p\times d}$. Since RiSE uses a single query, the class head $\phi_\mathrm{cls}$ can be ignored. The mask head maps them into mask-space features $\mathbf{Q}_\mathrm{mask}=\phi_{\mathrm{mlp}}(\mathbf{Q}_\mathrm{vfm})$ and $\mathbf{Z}_\mathrm{mask}=\phi_{\mathrm{up}}(\mathbf{Z}_\mathrm{vfm})$, where $\phi_{\mathrm{mlp}}$ is a three-layer MLP with GELU activations and $\phi_{\mathrm{up}}$ is a stack of blocks consisting of transposed-convolution, GELU, depth-wise convolution, and LayerNorm2d. The transformed query embedding and dense feature map are combined by the matching operation \texttt{torch.einsum("bqc,bchw->bqhw", Q\_mask, Z\_mask)}, followed by the sigmoid. 

\begin{center}
\begin{minipage}{0.92\linewidth}
\begin{lstlisting}[style=idepython,caption={Deterministic forward pass of the mask module.},label={lst:mask_module_forward}]
# Q_vfm: [B, N_q, d], Z_vfm: [B, Hp*Wp, d]
def mask_module_forward(Q_vfm, Z_vfm):
    Z_vfm = Z_vfm.transpose(1, 2).reshape(B, d, Hp, Wp)

    Q_mask = phi_mlp(Q_vfm)      # Linear-GELU-Linear-GELU-Linear
    Z_mask = phi_up(Z_vfm)       # repeated Conv-GELU-Conv-LayerNorm2d blocks
    mask_logits = torch.einsum("bqc,bchw->bqhw", Q_mask, Z_mask)
    probs = mask_logits.sigmoid()
    return probs
\end{lstlisting}
\end{minipage}
\end{center}

\paragraph{Moment Propagation.}
LPC perturbs only the dense feature $\mathbf{Z}_\mathrm{vfm}$ with Gaussian noise and the expected latent-perturbed output is
\[
\bar{\mathbf{p}}_{\mathrm{lat}}=\mathbb{E}_{\boldsymbol{\epsilon}\sim\mathcal{N}(\mathbf{0},\sigma^2 I)}\bigl[\phi_{\mathrm{mask}}(\mathbf{Q}_\mathrm{vfm}, \mathbf{Z}_\mathrm{vfm}+\boldsymbol{\epsilon})\bigr],
\]
where $\phi_{\mathrm{mask}}(\mathbf{Q}_\mathrm{vfm},\cdot)$ is the mask module with deterministic query input $\mathbf{Q}_\mathrm{vfm}$. We define $\phi_{\mathrm{mask}}$ as the composition of all layers in the mask module:
\[
\phi_{\mathrm{mask}} = \phi_{\mathrm{mask}}^{(L)} \circ \cdots \circ \phi_{\mathrm{mask}}^{(1)}.
\]
Since the nonlinear layers in $\phi_{\mathrm{mask}}$ do not preserve Gaussianity, the expectation is not available in closed form. We therefore apply ADF, which propagates a Gaussian approximation through the mask module. Let $\mathbf{Z}^{(0)}=\mathbf{Z}_\mathrm{vfm}+\boldsymbol{\epsilon}$ and
\[
\mathbf{Z}^{(\ell)} = \phi_{\mathrm{mask}}^{(\ell)}(\mathbf{Z}^{(\ell-1)}), \qquad \ell=1,\dots,L.
\]
ADF maintains at each layer
\[
q_\ell(\mathbf{Z}^{(\ell)})\approx \mathcal{N}(\boldsymbol{\mu}_\ell,\operatorname{diag}(\mathbf{v}_\ell)).
\]
Here $q_\ell(\mathbf{Z}^{(\ell)})$ denotes the ADF approximate distribution of the random intermediate feature $\mathbf{Z}^{(\ell)}$ at layer $\ell$. Starting from $q_0(\mathbf{Z}^{(0)})=\mathcal{N}(\mathbf{Z}_\mathrm{vfm},\sigma^2\mathbf{I})$, each layer takes a Gaussian input, applies $\phi_{\mathrm{mask}}^{(\ell)}$, and then re-approximates the resulting output by matching its mean and variance. The query projection $\mathbf{Q}_\mathrm{mask}=\phi_{\mathrm{mlp}}(\mathbf{Q}_\mathrm{vfm})$ is deterministic, so uncertainty is propagated only through $\phi_{\mathrm{up}}(\mathbf{Z}_\mathrm{vfm}+\boldsymbol{\epsilon})$ and the final matching operation. Tables~\ref{tab:adf_mean_prop} and~\ref{tab:adf_var_prop} summarize the corresponding mean and variance updates. In these tables, $(\boldsymbol{\mu}_x,\mathbf{v}_x)$ denotes the input mean and variance tensors, while $(\boldsymbol{\mu}_y,\mathbf{v}_y)$ denotes the output mean and variance tensors. The ADF computation is summarized with the PyTorch-style Pseudocode~\ref{lst:mask_module_agp}. After AGP produces the probability mean $\bar{\mathbf{p}}_\mathrm{lat}$, LPC computes SDC on $\bar{\mathbf{p}}_\mathrm{lat}$ for confidence. The final prediction is still taken from the original, unperturbed output.

\begin{table}[t]
\centering
\footnotesize
\setlength{\tabcolsep}{4pt}
\renewcommand{\arraystretch}{1.25}
\caption{ADF mean propagation rules in the mask module.}
\label{tab:adf_mean_prop}
\resizebox{\textwidth}{!}{%
\begin{tabular}{p{0.16\textwidth}p{0.4\textwidth}p{0.5\textwidth}}
    \toprule
    Operation & Formulation & Notes \\
    \midrule
    Convolution & $\boldsymbol{\mu}_y=\mathbf{W}*\boldsymbol{\mu}_x+\mathbf{b}$ & $\mathbf{W}$ is the convolution kernel and $*$ denotes convolution. The matching operation \texttt{torch.einsum("bqc,bchw->bqhw", Q\_mask, Z\_mask)} can be viewed as query-wise $1\times 1$ convolution with deterministic weights $\mathbf{Q}_\mathrm{mask}$. \\
    
    GELU & $\boldsymbol{\mu}_y\approx \frac{1}{\sqrt{\pi}}\sum_k w_k g(\sqrt{2\mathbf{v}_x}\,\xi_k + \boldsymbol{\mu}_x)$ & It is a Gauss--Hermite quadrature approximation under the Gaussian input, using nodes and weights $(\xi_k,w_k)$ with default $K=5$ and $g$ is GELU. \\
    
    LayerNorm2d & $\boldsymbol{\mu}_{y,c}=\frac{(\boldsymbol{\mu}_{x,c}-\bar{\boldsymbol{\mu}}_x)\gamma_c}{\sqrt{\mathbf{s}_{\mathrm{LN}}^2}}+\beta_c$ & $\bar{\boldsymbol{\mu}}_x=\frac{1}{d}\sum_c \boldsymbol{\mu}_{x,c}$ is the input channel mean, $\mathbf{s}_{\mathrm{LN}}^2\approx\operatorname{Var}_c(\boldsymbol{\mu}_{x,c})+(1-\frac{1}{d})\frac{1}{d}\sum_c\mathbf{v}_{x,c}$ is the expected channel variance and $\gamma_c,\beta_c$ are the LayerNorm affine parameters. \\
    
    Sigmoid & $\boldsymbol{\mu}_y\approx \frac{1}{2}\left[\sigma\!\left(\boldsymbol{\mu}_x-\sqrt{\mathbf{v}_x}\right)+\sigma\!\left(\boldsymbol{\mu}_x+\sqrt{\mathbf{v}_x}\right)\right]$ & This is the 2-point Gauss--Hermite quadrature for $\mathbb{E}[\sigma(z)]$ with $z\sim\mathcal{N}(\boldsymbol{\mu}_x,\mathbf{v}_x)$: after rescaling the Gaussian variable, the standard nodes become $\boldsymbol{\mu}_x\pm\sqrt{\mathbf{v}_x}$ and both have weight $\frac{1}{2}$. \\
    \bottomrule
\end{tabular}%
}
\end{table}

\begin{table}[t]
\centering
\footnotesize
\setlength{\tabcolsep}{4pt}
\renewcommand{\arraystretch}{1.25}
\caption{ADF variance propagation rules in the mask module.}
\label{tab:adf_var_prop}
\resizebox{\textwidth}{!}{%
\begin{tabular}{p{0.16\textwidth}p{0.4\textwidth}p{0.5\textwidth}}
    \toprule
    Operation & Formulation & Notes \\
    \midrule
    Convolution & $\mathbf{v}_y=(\mathbf{W}\odot\mathbf{W})*\mathbf{v}_x$ & $\odot$ denotes element-wise multiplication. The matching operation \texttt{torch.einsum("bqc,bchw->bqhw", Q\_mask, Z\_mask)} likewise follows this rule with squared deterministic query weights $\mathbf{Q}_\mathrm{mask}\odot\mathbf{Q}_\mathrm{mask}$, which yields the variance update used in Pseudocode~\ref{lst:mask_module_agp}. \\
    
    GELU & $\mathbf{v}_y\approx \frac{1}{\sqrt{\pi}}\sum_k w_k g(\sqrt{2\mathbf{v}_x}\,\xi_k + \boldsymbol{\mu}_x)^2-\boldsymbol{\mu}_y^2$ & $\boldsymbol{\mu}_y$ is the GELU mean approximation from the mean table. \\
    
    LayerNorm2d & $\mathbf{v}_{y,c}=\frac{\tilde{\mathbf{v}}_{x,c}\gamma_c^2}{\mathbf{s}_{\mathrm{LN}}^2}$ &  $\tilde{\mathbf{v}}_{x,c}=(1-\frac{2}{d})\mathbf{v}_{x,c}+\frac{1}{d^2}\sum_{c'}\mathbf{v}_{x,c'}$ is the variance after channel centering \\

    \bottomrule
\end{tabular}%
}
\end{table}

\begin{center}
\begin{minipage}{0.98\linewidth}
\begin{lstlisting}[style=idepython,caption={ADF computation in the mask module.},label={lst:mask_module_agp}]
# Q_vfm: [B, N_q, d], Z_vfm: [B, Hp*Wp, d]
def mask_module_agp(Q_vfm, Z_vfm, sigma):
    Z_vfm = Z_vfm.transpose(1, 2).reshape(B, d, Hp, Wp)

    mu = Z_vfm
    var = torch.full_like(mu, sigma ** 2)

    Q_mask = phi_mlp(Q_vfm)             # deterministic query projection
    mu_u, var_u = agp_phi_up(mu, var)   # Conv-GELU-Conv-LayerNorm2d AGP rules
    mu_m = torch.einsum("bqc,bchw->bqhw", Q_mask, mu_u)
    var_m = torch.einsum("bqc,bchw->bqhw", Q_mask.square(), var_u)

    mu_p = agp_sigmoid_mean(mu_m, var_m)
    return mu_p                         # mu_p is the LPC-averaged probability map
\end{lstlisting}
\end{minipage}
\end{center}

\captionsetup[table]{skip=8pt}
\section{Additional Experiments and Analyses}
\label{app:additional_experiments}

This section examines how the learned embeddings partition forgery types, whether those partitions need type metadata, and how individual experts behave across datasets. We then study routing, transfer to natural images, robustness to image transformations, and inference latency. Unless otherwise stated, F1 is the localization metric. 

\subsection{Cluster Quality}
\label{app:cluster_results}

\begin{figure}[tbp]
\centering
\includegraphics[width=\linewidth]{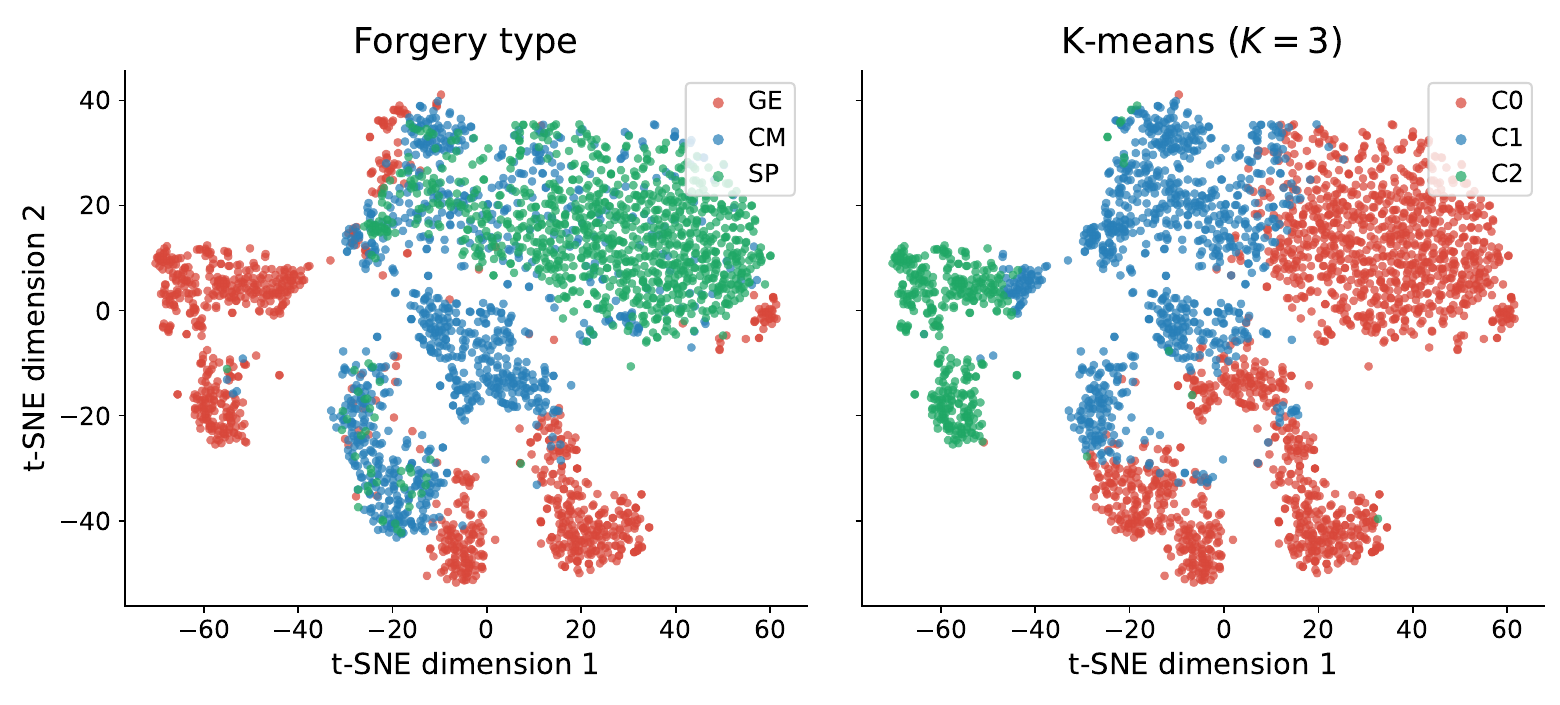}
\caption{RiSE-1 query embeddings for 3{,}000 balanced training images. The same points are colored by forgery type (left) and $K=3$ cluster (right).}
\label{fig:cluster_scatter_k3}
\end{figure}

\begin{figure}[tbp]
\centering
\includegraphics[width=\linewidth]{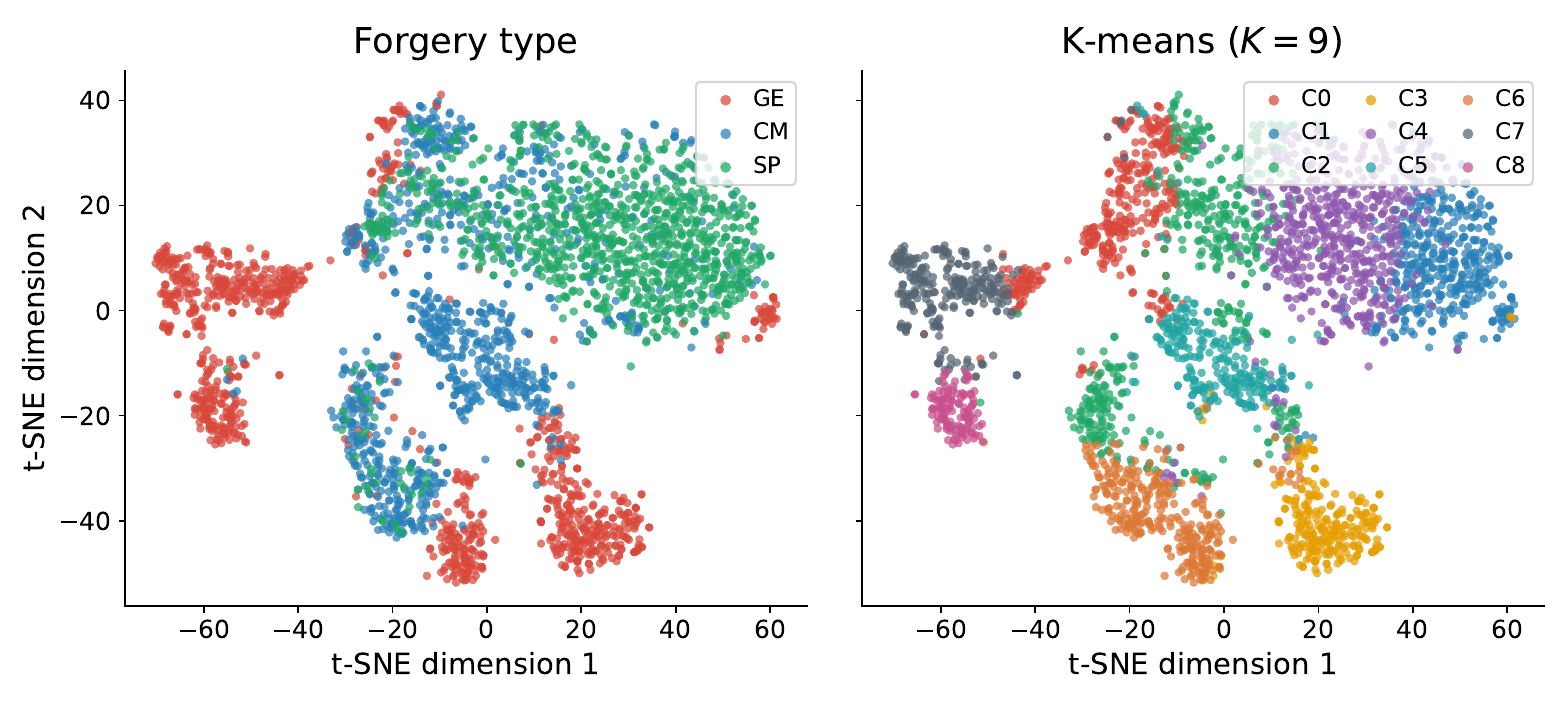}
\caption{The same embeddings colored by forgery type (left) and $K=9$ cluster (right).}
\label{fig:cluster_scatter_k9}
\end{figure}

We assess whether RiSE-1 query embeddings contain forgery-related structure using 3{,}000 DocTamper training images, with 1{,}000 images sampled from each of the generation (GE), copy-move (CM), and splicing (SP) categories. K-means is applied to the query embeddings, and the forgery types color the same points in the left panels of Figs.~\ref{fig:cluster_scatter_k3} and~\ref{fig:cluster_scatter_k9}. The t-SNE projection is used only for visualization. Overall, the embeddings show a clear cluster between the three forgery types. 
With a simply K-mean partition, the clusters broadly align with the forgery types, but there is also overlap between clusters and forgery types. The right panels of Figs.~\ref{fig:cluster_scatter_k3} and~\ref{fig:cluster_scatter_k9} show the same embeddings colored by cluster assignment for $K=3$ and $K=9$, respectively. We also report silhouette scores over five random seeds to show the stability of the clustering results in Table~\ref{tab:cluster_silhouette}.

\begin{table}[t]
\centering
\footnotesize
\caption{Silhouette scores of the embedding clusters over five random seeds (mean $\pm$ standard deviation).}
\label{tab:cluster_silhouette}
\begin{tabular}{lcc}
\toprule
Metric & $K=3$ & $K=9$ \\
\midrule
Silhouette score & $0.1232 \pm 0.0000$ & $0.0789 \pm 0.0001$ \\
\bottomrule
\end{tabular}
\end{table}

\subsection{Metadata-Free Partitioning}

\begin{table}[tbp]
\centering
\footnotesize
\setlength{\tabcolsep}{6pt}
\caption{RiSE-3 F1 with metadata-based and metadata-free partitions.}
\label{tab:metadata_free_performance}
\rowcolors{2}{TableStripe}{white}
\begin{tabular}{lcccc}
\toprule
Dataset & Metadata-based / LPC & Metadata-based / oracle & Metadata-free / LPC & Metadata-free / oracle \\
\midrule
FCD & \scorecell{83}{0.710} & \scorecell{89}{0.746} & \scorecell{92}{0.771} & \scorecell{97}{0.801} \\
SCD & \scorecell{73}{0.643} & \scorecell{83}{0.707} & \scorecell{68}{0.614} & \scorecell{77}{0.669} \\
Test & \scorecell{85}{0.721} & \scorecell{90}{0.757} & \scorecell{79}{0.683} & \scorecell{85}{0.723} \\
AvgI & \scorecell{80}{0.691} & \scorecell{87}{0.737} & \scorecell{80}{0.689} & \scorecell{86}{0.731} \\
\midrule
T-SROIE & \scorecell{91}{0.760} & \scorecell{99}{0.815} & \scorecell{94}{0.783} & \scorecell{100}{0.820} \\
OSTF & \scorecell{70}{0.625} & \scorecell{75}{0.658} & \scorecell{64}{0.588} & \scorecell{68}{0.615} \\
TPIC-13 & \scorecell{88}{0.740} & \scorecell{92}{0.766} & \scorecell{89}{0.747} & \scorecell{91}{0.764} \\
RTM & \scorecell{0}{0.173} & \scorecell{7}{0.221} & \scorecell{0}{0.173} & \scorecell{6}{0.213} \\
AvgC & \scorecell{62}{0.575} & \scorecell{68}{0.615} & \scorecell{62}{0.573} & \scorecell{66}{0.603} \\
\midrule
Avg. all & \scorecell{70}{0.625} & \scorecell{76}{0.667} & \scorecell{70}{0.623} & \scorecell{75}{0.658} \\
\bottomrule
\end{tabular}
\end{table}

Since the RiSE-3 require forgery type metadata to assign experts, we investigate whether the expert partition can be learned without metadata. Table~\ref{tab:metadata_free_performance} compares the two partitions of RiSE-3 with both LPC selection and oracle selection. In the metadata-based partition, we use the three ground-truth forgery types, while the metadata-free partition uses the K-means ($K=3$) clusters. The overall F1 of the two partitions is similar, with the metadata-free partition slightly outperforming the metadata-based one. The results suggest that the learned embeddings can be used to partition the training set without requiring metadata, which is useful for datasets where forgery type labels are unavailable.

\subsection{Per-Expert Performance}
\label{app:expert_routing_results}

\begin{table}[tbp]
\centering
\footnotesize
\setlength{\tabcolsep}{4pt}
\caption{Per-expert F1 for RiSE-3. Each row is one expert trained on the generation (GE), copy-move (CM), or splicing (SP) subset. Dataset order matches Table~\ref{tab:rise9_expert_f1}}
\label{tab:rise3_expert_f1}
\rowcolors{2}{TableStripe}{white}
\begin{tabular}{lcccccccc}
\toprule
Expert & Test & FCD & SCD & T-SROIE & TPIC-13 & RTM & OSTF & Avg. \\
\midrule
$E_{\rm GE}$ & \scorecell{43}{0.393} & \scorecell{75}{0.600} & \scorecell{51}{0.445} & \scorecell{97}{0.743} & \scorecell{100}{0.763} & \scorecell{0}{0.113} & \scorecell{80}{0.631} & \scorecell{64}{0.527} \\
$E_{\rm CM}$ & \scorecell{76}{0.605} & \scorecell{80}{0.634} & \scorecell{68}{0.556} & \scorecell{76}{0.605} & \scorecell{18}{0.231} & \scorecell{5}{0.148} & \scorecell{18}{0.227} & \scorecell{49}{0.430} \\
$E_{\rm SP}$ & \scorecell{70}{0.571} & \scorecell{86}{0.669} & \scorecell{57}{0.483} & \scorecell{62}{0.519} & \scorecell{21}{0.248} & \scorecell{0}{0.114} & \scorecell{27}{0.286} & \scorecell{46}{0.413} \\
\bottomrule
\end{tabular}
\end{table}

\begin{table}[htbp]
\centering
\footnotesize
\setlength{\tabcolsep}{3pt}
\caption{Per-expert F1 scores for RiSE-9. Each row is one expert $E_{t,j}$ for forgery type $t\in\{\rm GE,CM,SP\}$ and within-type cluster $j\in\{0,1,2\}$}
\label{tab:rise9_expert_f1}
\rowcolors{2}{TableStripe}{white}
\begin{tabular}{lcccccccc}
\toprule
Expert & Test & FCD & SCD & T-SROIE & TPIC-13 & RTM & OSTF & Avg. \\
\midrule
$E_{\rm GE,0}$ & \scorecell{36}{0.335} & \scorecell{58}{0.496} & \scorecell{49}{0.433} & \scorecell{98}{0.783} & \scorecell{83}{0.672} & \scorecell{2}{0.089} & \scorecell{64}{0.542} & \scorecell{56}{0.479} \\
$E_{\rm GE,1}$ & \scorecell{23}{0.244} & \scorecell{52}{0.449} & \scorecell{16}{0.195} & \scorecell{61}{0.519} & \scorecell{86}{0.698} & \scorecell{3}{0.096} & \scorecell{70}{0.580} & \scorecell{44}{0.397} \\
$E_{\rm GE,2}$ & \scorecell{46}{0.407} & \scorecell{72}{0.594} & \scorecell{58}{0.494} & \scorecell{93}{0.749} & \scorecell{100}{0.798} & \scorecell{6}{0.122} & \scorecell{84}{0.683} & \scorecell{66}{0.550} \\
$E_{\rm CM,0}$ & \scorecell{60}{0.513} & \scorecell{66}{0.553} & \scorecell{59}{0.505} & \scorecell{40}{0.365} & \scorecell{15}{0.182} & \scorecell{5}{0.113} & \scorecell{20}{0.222} & \scorecell{38}{0.350} \\
$E_{\rm CM,1}$ & \scorecell{71}{0.587} & \scorecell{90}{0.727} & \scorecell{54}{0.466} & \scorecell{72}{0.598} & \scorecell{19}{0.216} & \scorecell{8}{0.136} & \scorecell{22}{0.237} & \scorecell{48}{0.424} \\
$E_{\rm CM,2}$ & \scorecell{52}{0.452} & \scorecell{56}{0.480} & \scorecell{61}{0.520} & \scorecell{77}{0.630} & \scorecell{14}{0.176} & \scorecell{6}{0.117} & \scorecell{9}{0.140} & \scorecell{39}{0.359} \\
$E_{\rm SP,0}$ & \scorecell{60}{0.513} & \scorecell{60}{0.512} & \scorecell{52}{0.451} & \scorecell{57}{0.487} & \scorecell{7}{0.126} & \scorecell{0}{0.080} & \scorecell{11}{0.153} & \scorecell{35}{0.332} \\
$E_{\rm SP,1}$ & \scorecell{71}{0.586} & \scorecell{83}{0.679} & \scorecell{58}{0.498} & \scorecell{66}{0.554} & \scorecell{25}{0.259} & \scorecell{6}{0.120} & \scorecell{28}{0.276} & \scorecell{48}{0.425} \\
$E_{\rm SP,2}$ & \scorecell{59}{0.501} & \scorecell{55}{0.477} & \scorecell{50}{0.434} & \scorecell{52}{0.450} & \scorecell{7}{0.128} & \scorecell{0}{0.077} & \scorecell{12}{0.161} & \scorecell{33}{0.318} \\
\bottomrule
\end{tabular}
\end{table}

Tables~\ref{tab:rise3_expert_f1} and~\ref{tab:rise9_expert_f1} show the per-expert F1 scores for RiSE-3 and RiSE-9. RiSE-3 experts $E_{\rm GE}$, $E_{\rm CM}$, and $E_{\rm SP}$ are trained on the corresponding forgery-type subsets. RiSE-9 splits each forgery subset into three smaller clusters, yielding $E_{t,j}$ for type $t$ and cluster $j$. The large performance variation among experts shows that different experts capture distinct forgery features and document semantics, which directly aligns with our expert-scaling motivation. For example, RiSE-3 $E_{\rm GE}$ is strongest on T-SROIE, TPIC-13, and OSTF, whereas $E_{\rm CM}$ and $E_{\rm SP}$ perform better on several in-domain sets and RTM. The RiSE average F1 exceeds that of every individual expert, reaching $0.625$ for RiSE-3 and $0.650$ for RiSE-9, while remaining close to the oracle scores of $0.667$ and $0.723$. RTM remains difficult for every expert and even for the oracle, which confirms that the main limitation is insufficient manual-forgery expertise.

\subsection{LPC Routing Accuracy}

\begin{table}[htbp]
\centering
\footnotesize
\setlength{\tabcolsep}{5pt}
\caption{LPC routing accuracy against the oracle-selected expert.}
\label{tab:lpc_routing_accuracy}
\rowcolors{2}{TableStripe}{white}
\begin{tabular}{lcccc}
\toprule
Dataset & RiSE-3 Top-1 & RiSE-9 Top-1 & RiSE-9 Top-3 & RiSE-9 Top-5 \\
\midrule
Test & \scorecell{53}{0.620} & \scorecell{35}{0.483} & \scorecell{70}{0.754} & \scorecell{88}{0.887} \\
FCD & \scorecell{60}{0.678} & \scorecell{56}{0.644} & \scorecell{84}{0.856} & \scorecell{95}{0.938} \\
SCD & \scorecell{46}{0.572} & \scorecell{12}{0.311} & \scorecell{63}{0.695} & \scorecell{83}{0.851} \\
T-SROIE & \scorecell{54}{0.630} & \scorecell{17}{0.346} & \scorecell{68}{0.734} & \scorecell{88}{0.889} \\
TPIC-13 & \scorecell{84}{0.855} & \scorecell{43}{0.544} & \scorecell{96}{0.946} & \scorecell{100}{0.978} \\
RTM & \scorecell{27}{0.425} & \scorecell{0}{0.221} & \scorecell{36}{0.491} & \scorecell{62}{0.689} \\
OSTF & \scorecell{73}{0.776} & \scorecell{37}{0.501} & \scorecell{88}{0.889} & \scorecell{97}{0.957} \\
Average & \scorecell{57}{0.651} & \scorecell{28}{0.435} & \scorecell{72}{0.766} & \scorecell{88}{0.884} \\
\bottomrule
\end{tabular}
\end{table}

\begin{figure}[tbp]
\centering
\begin{subfigure}[t]{0.49\linewidth}
\centering
\includegraphics[width=\linewidth]{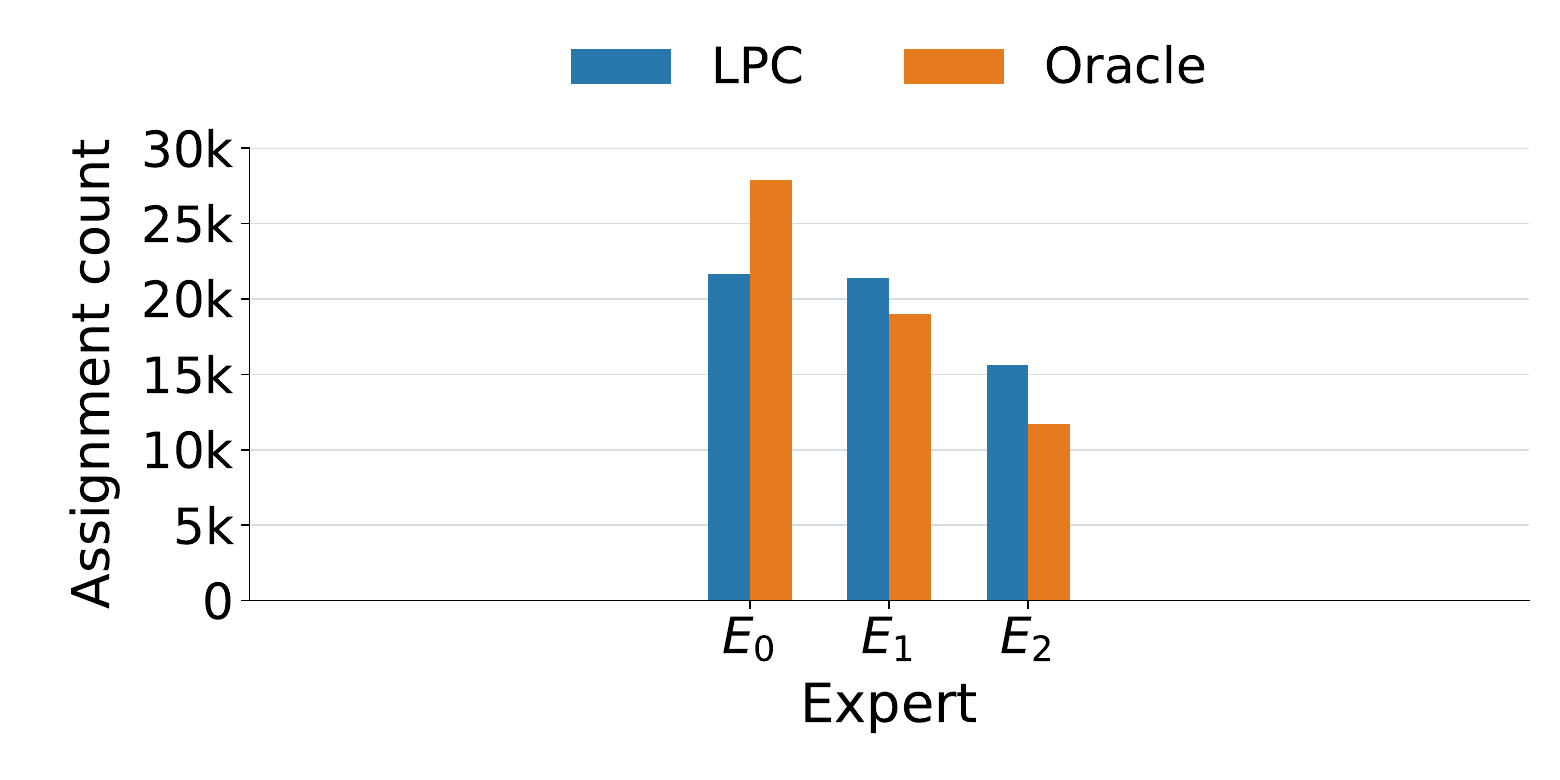}
\caption{RiSE-3}
\label{fig:rise3_routing_counts}
\end{subfigure}\hfill
\begin{subfigure}[t]{0.49\linewidth}
\centering
\includegraphics[width=\linewidth]{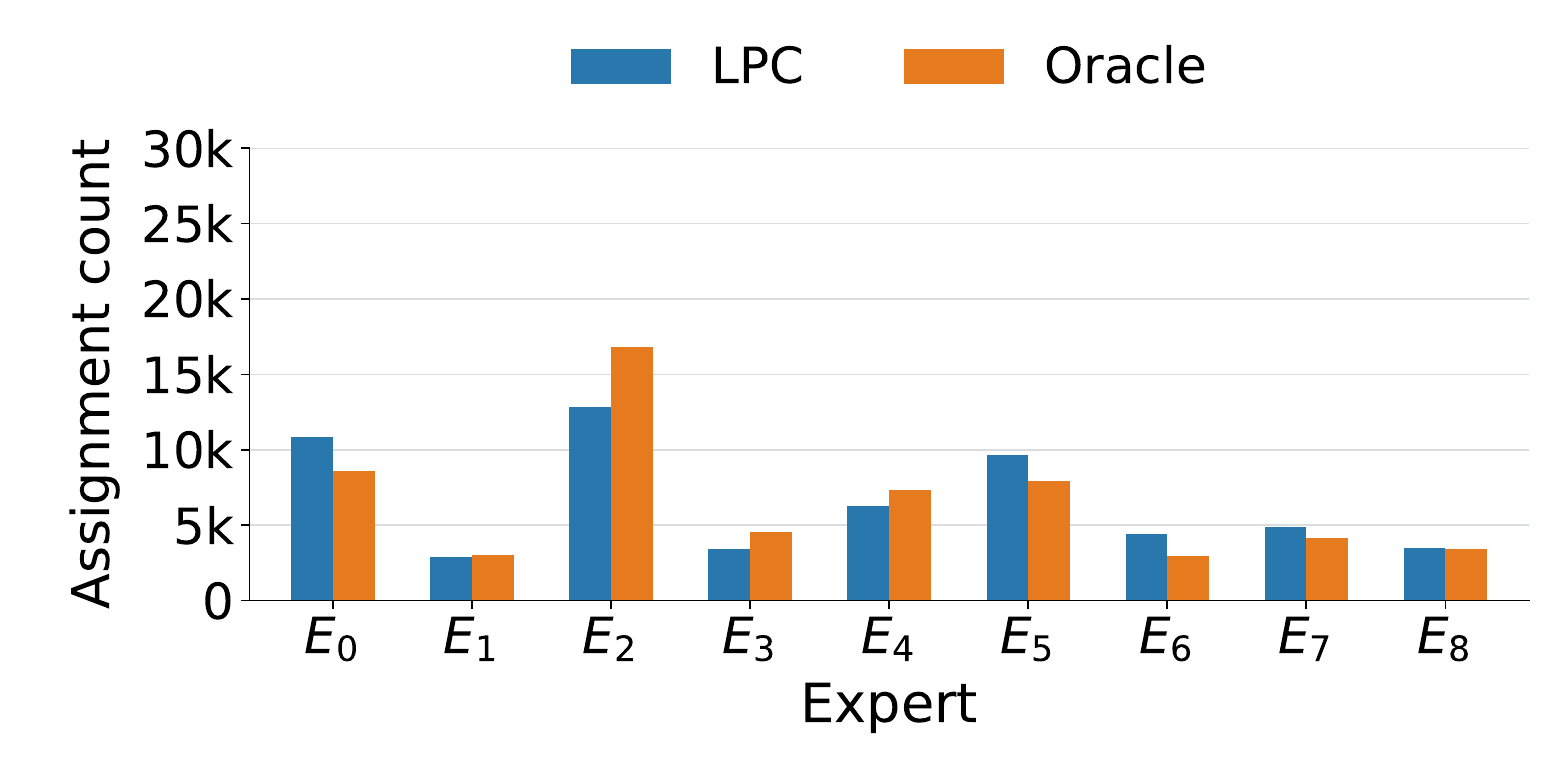}
\caption{RiSE-9}
\label{fig:rise9_routing_counts}
\end{subfigure}
\caption{Expert selection counts across seven test sets: LPC assignments and ground-truth oracle choices for (a) RiSE-3 and (b) RiSE-9.}
\label{fig:routing_counts}
\end{figure}

We measure LPC routing accuracy against the oracle-selected expert. For image $\mathbf{X}_i$, let $j_i^*$ denote the oracle expert index, and let $\mathcal{R}_i^{\mathrm{LPC}}(k)$ denote the index set of $k$ experts with the highest LPC confidence. For $N$ images, we define
\[
\mathrm{LPC\ Top\text{-}}k\ \mathrm{ACC}
= \frac{1}{N}\sum_{i=1}^{N}
\mathds{1}\!\left[j_i^* \in \mathcal{R}_i^{\mathrm{LPC}}(k)\right].
\]
The oracle uses the ground-truth mask to choose the best expert for each image, so it is an unreachable reference for deployment. Table~\ref{tab:lpc_routing_accuracy} reports the routing accuracy for RiSE-3 and RiSE-9. For RiSE-9, the oracle expert is ranked within the top-3 candidates with $0.766$ average accuracy and within the top-5 with $0.884$ average accuracy. This indicates that LPC retains useful expert preference information despite the harder 9-expert selection problem. RTM remains the most difficult domain, with $0.221$ Top-1 and $0.689$ Top-5 accuracy, consistent with its manual manipulations being hard to represent by the source-domain experts. Fig.~\ref{fig:routing_counts} compares the aggregate assignment counts; LPC and the oracle distribute images differently among experts. These counts describe routing frequency, whereas exact expert agreement and end-to-end localization performance measure different outcomes.

\subsection{Zero-Shot Transfer to Natural-Image Forgeries}

\begin{table}[htbp]
\centering
\footnotesize
\setlength{\tabcolsep}{5pt}
\caption{Zero-shot F1 on six natural-image forgery datasets. All methods are trained only on DocTamper.}
\label{tab:natural_image_transfer}
\rowcolors{2}{TableStripe}{white}
\begin{tabular}{lccccccc}
\toprule
Method & NIST16 & Columbia & DSO & CASIA & CASIA2 & IMD2020 & Avg. \\
\midrule
DTD & \scorecell{21}{0.059} & \scorecell{71}{0.198} & \scorecell{41}{0.116} & \scorecell{0}{0.001} & \scorecell{5}{0.014} & \scorecell{27}{0.076} & \scorecell{27}{0.077} \\
FFDN & \scorecell{9}{0.026} & \scorecell{5}{0.016} & \scorecell{0}{0.001} & \scorecell{2}{0.006} & \scorecell{3}{0.010} & \scorecell{6}{0.017} & \scorecell{4}{0.013} \\
ADCD-Net & \scorecell{37}{0.103} & \scorecell{10}{0.030} & \scorecell{16}{0.047} & \scorecell{6}{0.019} & \scorecell{13}{0.038} & \scorecell{21}{0.060} & \scorecell{17}{0.049} \\
RiSE-1 & \scorecell{18}{0.052} & \scorecell{1}{0.003} & \scorecell{1}{0.005} & \scorecell{21}{0.060} & \scorecell{16}{0.046} & \scorecell{16}{0.045} & \scorecell{12}{0.035} \\
RiSE-3 & \scorecell{42}{0.117} & \scorecell{11}{0.031} & \scorecell{9}{0.026} & \scorecell{50}{0.141} & \scorecell{39}{0.110} & \scorecell{55}{0.155} & \scorecell{34}{0.097} \\
RiSE-9 & \scorecell{77}{0.216} & \scorecell{25}{0.072} & \scorecell{10}{0.030} & \scorecell{100}{0.280} & \scorecell{53}{0.148} & \scorecell{68}{0.190} & \scorecell{56}{0.156} \\
\bottomrule
\end{tabular}
\end{table}

All methods in Table~\ref{tab:natural_image_transfer} are trained on DocTamper only and evaluated without target-domain adaptation on six natural-image forgery datasets. Absolute F1 is low under this substantial shift from text-on-white documents to natural scenes, people, animals, and objects. RiSE-9 achieves the highest mean F1 of 0.156, compared with 0.077 for the strongest baseline, DTD.

\subsection{Robustness to JPEG Recompression and Resizing}

\begin{table}[htbp]
\centering
\footnotesize
\setlength{\tabcolsep}{3pt}
\caption{Robustness under JPEG recompression (J90, J80, J70: quality factors) and resizing (R90, R80, R70: retained linear scale in percent). Avg. averages the six transformed conditions; Clean is reported separately.}
\label{tab:rebuttal_robustness}
\rowcolors{2}{TableStripe}{white}
\begin{tabular}{llcccccccc}
\toprule
Dataset & Method & Clean & J90 & J80 & J70 & R90 & R80 & R70 & Avg. \\
\midrule
\emph{T-SROIE} & ADCD-Net & \scorecell{59}{0.623} & \scorecell{47}{0.571} & \scorecell{38}{0.538} & \scorecell{31}{0.508} & \scorecell{92}{0.757} & \scorecell{90}{0.748} & \scorecell{92}{0.755} & \scorecell{65}{0.646} \\
 & RiSE-9-DCT & \scorecell{85}{0.729} & \scorecell{73}{0.679} & \scorecell{63}{0.638} & \scorecell{56}{0.608} & \scorecell{97}{0.774} & \scorecell{98}{0.780} & \scorecell{91}{0.750} & \scorecell{80}{0.705} \\
 & RiSE-9 & \scorecell{100}{0.786} & \scorecell{78}{0.697} & \scorecell{68}{0.660} & \scorecell{61}{0.631} & \scorecell{94}{0.762} & \scorecell{96}{0.772} & \scorecell{93}{0.759} & \scorecell{82}{0.713} \\
\midrule
\emph{TPIC-13} & ADCD-Net & \scorecell{49}{0.579} & \scorecell{53}{0.599} & \scorecell{29}{0.499} & \scorecell{13}{0.436} & \scorecell{33}{0.516} & \scorecell{33}{0.515} & \scorecell{32}{0.513} & \scorecell{32}{0.513} \\
 & RiSE-9-DCT & \scorecell{82}{0.713} & \scorecell{82}{0.714} & \scorecell{60}{0.625} & \scorecell{43}{0.555} & \scorecell{77}{0.693} & \scorecell{74}{0.682} & \scorecell{72}{0.676} & \scorecell{68}{0.658} \\
 & RiSE-9 & \scorecell{100}{0.788} & \scorecell{97}{0.777} & \scorecell{84}{0.722} & \scorecell{67}{0.655} & \scorecell{95}{0.768} & \scorecell{93}{0.761} & \scorecell{93}{0.759} & \scorecell{88}{0.740} \\
\midrule
\emph{OSTF} & ADCD-Net & \scorecell{15}{0.441} & \scorecell{13}{0.434} & \scorecell{7}{0.411} & \scorecell{0}{0.382} & \scorecell{3}{0.396} & \scorecell{4}{0.397} & \scorecell{2}{0.390} & \scorecell{5}{0.402} \\
 & RiSE-9-DCT & \scorecell{53}{0.597} & \scorecell{61}{0.628} & \scorecell{56}{0.609} & \scorecell{49}{0.580} & \scorecell{56}{0.609} & \scorecell{55}{0.606} & \scorecell{56}{0.609} & \scorecell{55}{0.607} \\
 & RiSE-9 & \scorecell{75}{0.688} & \scorecell{73}{0.679} & \scorecell{70}{0.667} & \scorecell{65}{0.644} & \scorecell{70}{0.666} & \scorecell{69}{0.663} & \scorecell{70}{0.667} & \scorecell{69}{0.664} \\
\bottomrule
\end{tabular}
\end{table}

We additionally evaluate the robustness of RiSE-9 and RiSE-9-DCT on the images of T-SROIE, TPIC-13, and OSTF under JPEG recompression and resizing against ADCD-Net. ADCD-Net is a strong robust baseline specifically designed for forgery localization under various distortions including JPEG, resize, and crop. Table~\ref{tab:rebuttal_robustness} reports the F1 scores, and Avg. across all distortions is reported. RiSE-9 achieves the highest average F1 on all three datasets and remains robust under JPEG recompression and resizing. Relative to ADCD-Net, RiSE-9 improves the average F1 by $10.4\%$ on T-SROIE, $44.2\%$ on TPIC-13, and $65.2\%$ on OSTF. RiSE-9-DCT also consistently outperforms ADCD-Net, showing strong robustness when the DCT pathway is included with our DCT dropout strategy.

\subsection{Single-GPU Inference Latency}

To quantify the practical trade-off, we measure single-image latency for $512\times512$ inputs on a single RTX 4090. Table~\ref{tab:single_gpu_latency} reports the mean and standard deviation across 100 samples, together with the cross-domain AvgC F1. Each RiSE configuration is Pareto-efficient in the latency--AvgC plane. RiSE-1 is within $6.7$ ms of DTD and $2.3$ ms of FFDN, yet improves AvgC by $0.295$ and $0.242$, respectively. RiSE-3 strictly dominates ADCD-Net, with both lower latency ($0.1165$ vs. $0.1171$ s) and higher AvgC ($0.575$ vs. $0.450$). RiSE-9 is the accuracy-prioritized Pareto-efficient option. All $9$ experts run in parallel, fit on one RTX 4090 using 21{,}691 MiB, and process a $512\times512$ image in $0.3571$ s to achieve the highest AvgC of $0.611$. This suits server-side document verification, such as banking/KYC, where a sub-second decision is practical and missing a forgery is substantially more costly than the additional latency.

\begin{table}[htbp]
\centering
\footnotesize
\setlength{\tabcolsep}{5pt}
\caption{Cross-domain AvgC F1 and single-image latency on one NVIDIA RTX 4090. Latency is reported as mean $\pm$ standard deviation over 100 inputs.}
\label{tab:single_gpu_latency}
\rowcolors{2}{TableStripe}{white}
\begin{tabular}{lcc}
\toprule
Model & AvgC F1 & Latency (s) \\
\midrule
DTD & \scorecell{0}{0.247} & $0.0407 \pm 0.0006$ \\
FFDN & \scorecell{15}{0.300} & $0.0451 \pm 0.0009$ \\
RiSE-1 & \scorecell{81}{0.542} & $0.0474 \pm 0.0002$ \\
ADCD-Net & \scorecell{56}{0.450} & $0.1171 \pm 0.0001$ \\
RiSE-3 & \scorecell{90}{0.575} & $0.1165 \pm 0.0004$ \\
RiSE-9 & \scorecell{100}{0.611} & $0.3571 \pm 0.0005$ \\
\bottomrule
\end{tabular}
\end{table}

\section{Efficiency}

Table~\ref{tab:efficiency} reports the measured efficiency of the RiSE variants on $4$ NVIDIA RTX 4090 GPUs.

\paragraph{Parameter count.}
Relative to RiSE-1, scaling to RiSE-3 and RiSE-9 increases the total parameter count by only $1.003\times$ and $1.011\times$, while the trainable parameter count increases by $1.07\times$ and $1.29\times$. This mild growth is consistent with the RiSE design, where the dominant VFM parameters and the mask module are shared, and each additional expert contributes only a small residual subspace.

\paragraph{Training time.}
We adopt an MoE-style implementation so that multiple RiSE experts are optimized within one unified training process rather than launched as fully separate training jobs. Unlike conventional MoE, RiSE does not introduce a learnable router during training. Instead, the routing path is deterministically given by the expert-assignment label for each sample described in Section~\ref{app:domain_label_acquisition}. This fixed-routing implementation keeps the experts specialized on their assigned training subsets while reducing the overhead compared with training each expert completely in isolation. The main extra cost appears during training because more experts still require more forward and optimization runs, increasing training time by $2.66\times$ for RiSE-3 and $7.08\times$ for RiSE-9 relative to RiSE-1. Nevertheless, RiSE remains efficient in the training phase because each expert is trained for only 10k steps, which is $10\times$ fewer than the $100$k steps commonly used by existing TFL detectors \citep{du2025forensichub}. On the same $4$ RTX 4090 GPUs, the strong TFL baseline ADCD-Net requires approximately $27$ training hours, so even RiSE-9 is still about $3.4\times$ faster to train.

\paragraph{Inference time.}
For inference, we convert throughput to per-sample latency in seconds. Since the experts can be executed in parallel, inference time does not grow linearly with the number of experts. Compared with RiSE-1, the latency increases by only $1.61\times$ for RiSE-3 and $4.23\times$ for RiSE-9, even though the number of experts increases by $3\times$ and $9\times$, respectively. RiSE-9 still remains below $0.1$ seconds per sample on the same $4$-GPU setup.

\begin{table}[ht]
    \centering
    \footnotesize
    \setlength{\tabcolsep}{8pt}
    \renewcommand{\arraystretch}{1.15}
    \caption{Efficiency comparison of non-DCT RiSE variants measured on $4$ NVIDIA RTX 4090 GPUs. Training time is reported in hours; inference time is converted from samples-per-second throughput to seconds per sample.}
    \label{tab:efficiency}
    \resizebox{\linewidth}{!}{%
    \rowcolors{2}{TableStripe}{white}
    \begin{tabular}{lcccc}
        \toprule
        Method & \#Total Params (M) & \#Trainable Params (M) & Training Time (hour) & Inference Time (s/sample) \\
        \midrule
        RiSE-1 & 315.14 & 12.20 & 1.12 & 0.0219 \\
        RiSE-3 & 316.02 & 13.09 & 2.98 & 0.0353 \\
        RiSE-9 & 318.68 & 15.75 & 7.93 & 0.0927 \\
        \bottomrule
    \end{tabular}
    }
\end{table}

\subsection{Lightweight Expert Selection}
\label{app:lightweight_expert_selection}

\paragraph{Server-side forensic analysis.}
Inference cost increases with expert count, but our target TFL deployment is server-side forensic analysis rather than real-time edge inference. This setting can use powerful GPU resources and does not require real-time processing. It is consistent with practical document-verification scenarios such as online banking and Know Your Customer (KYC), where images are submitted for automated forensic analysis. Recent work also frames ID-document forgery detection and localization in online authentication settings~\citep{korshunov2025deepid}. 

\paragraph{Dynamic early exit.}
A dynamic early-exit mechanism is a natural extension. EoMT could decode segmentation predictions from intermediate ViT blocks with homogeneous feature shapes, allowing LPC-based confidence evaluation at intermediate layers for early termination, analogous to confident layer decoding in LLMs~\citep{zhang2026deeper}. Developing and evaluating such a strategy is beyond the scope of this paper, and we leave it for future work.

\paragraph{Expert merging.}
Merging expert models into a unified model is another promising direction. Existing work on task arithmetic, common and task-specific subspaces, and SVD-based alignment for LoRA merging could provide practical starting points~\citep{ilharco2023editing,marczak2025notask,stoica2025knots}. Adapting these methods to RiSE's residual subspace experts and evaluating the resulting accuracy--efficiency trade-off remain future work.

\section{Visualization}

\begin{figure}[t]
    \centering
    \begin{subfigure}[t]{0.32\linewidth}
        \centering
        \includegraphics[width=\linewidth]{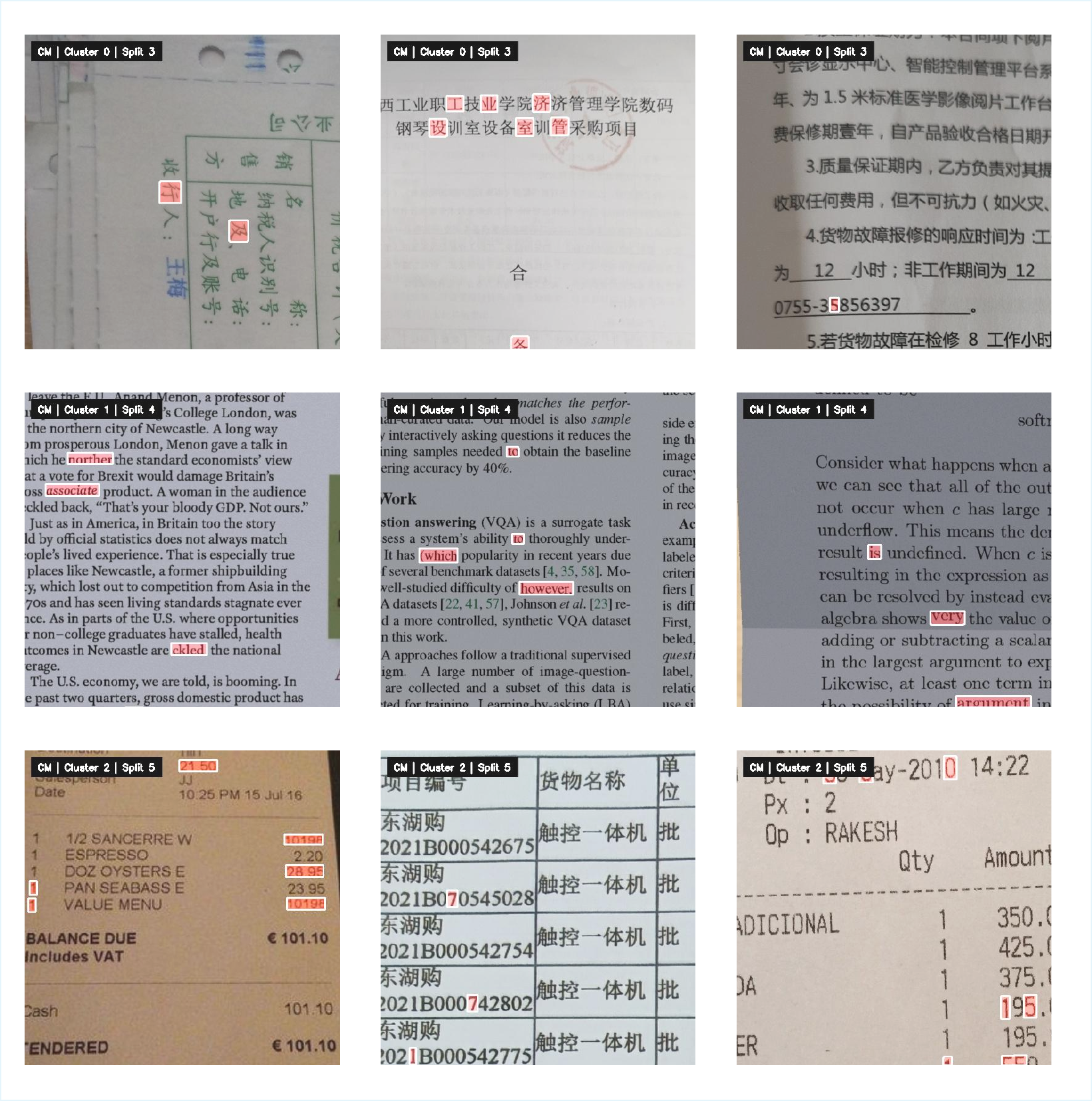}
        \caption{Copy-move}
    \end{subfigure}
    \hfill
    \begin{subfigure}[t]{0.32\linewidth}
        \centering
        \includegraphics[width=\linewidth]{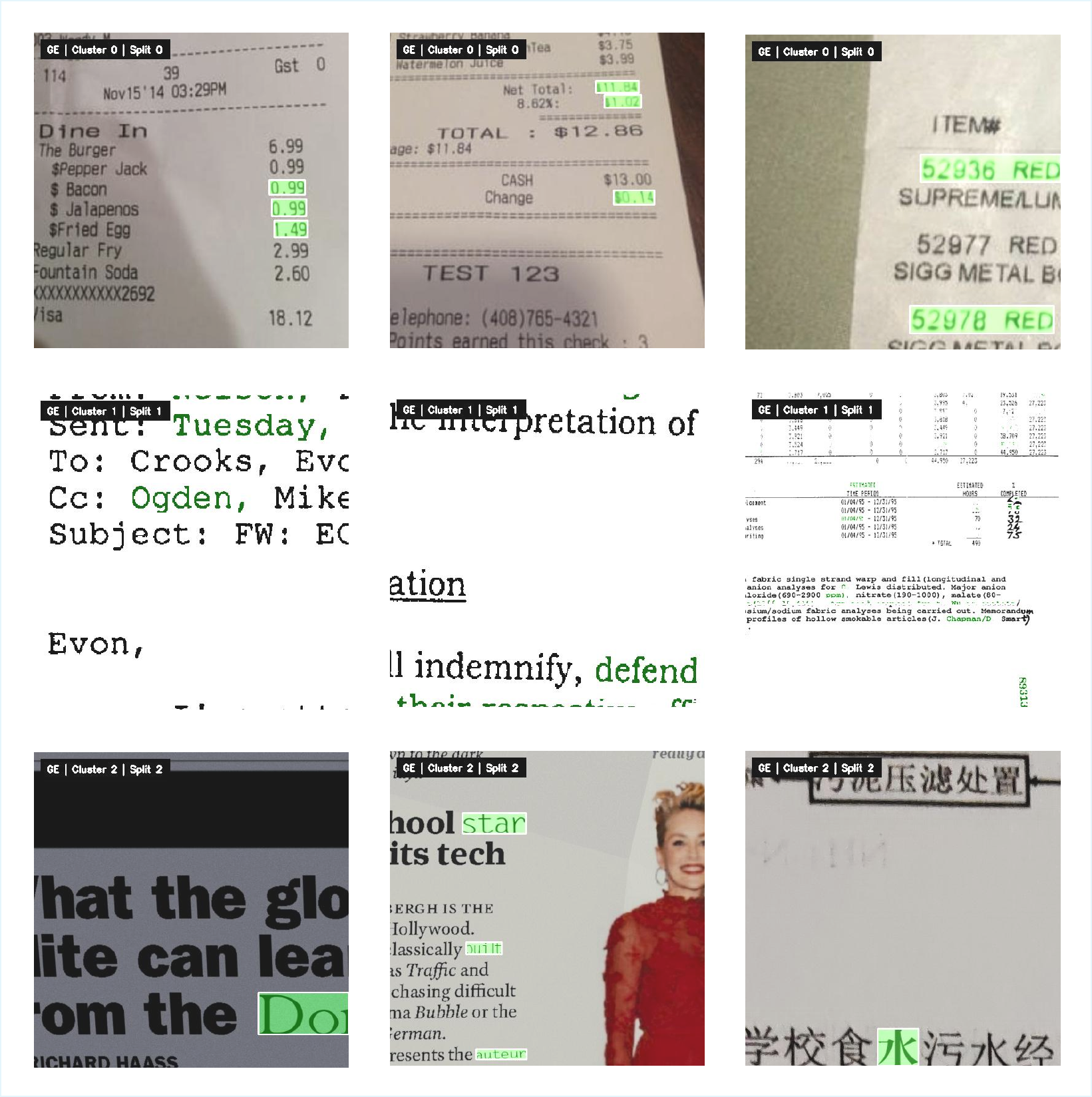}
        \caption{Generation}
    \end{subfigure}
    \hfill
    \begin{subfigure}[t]{0.32\linewidth}
        \centering
        \includegraphics[width=\linewidth]{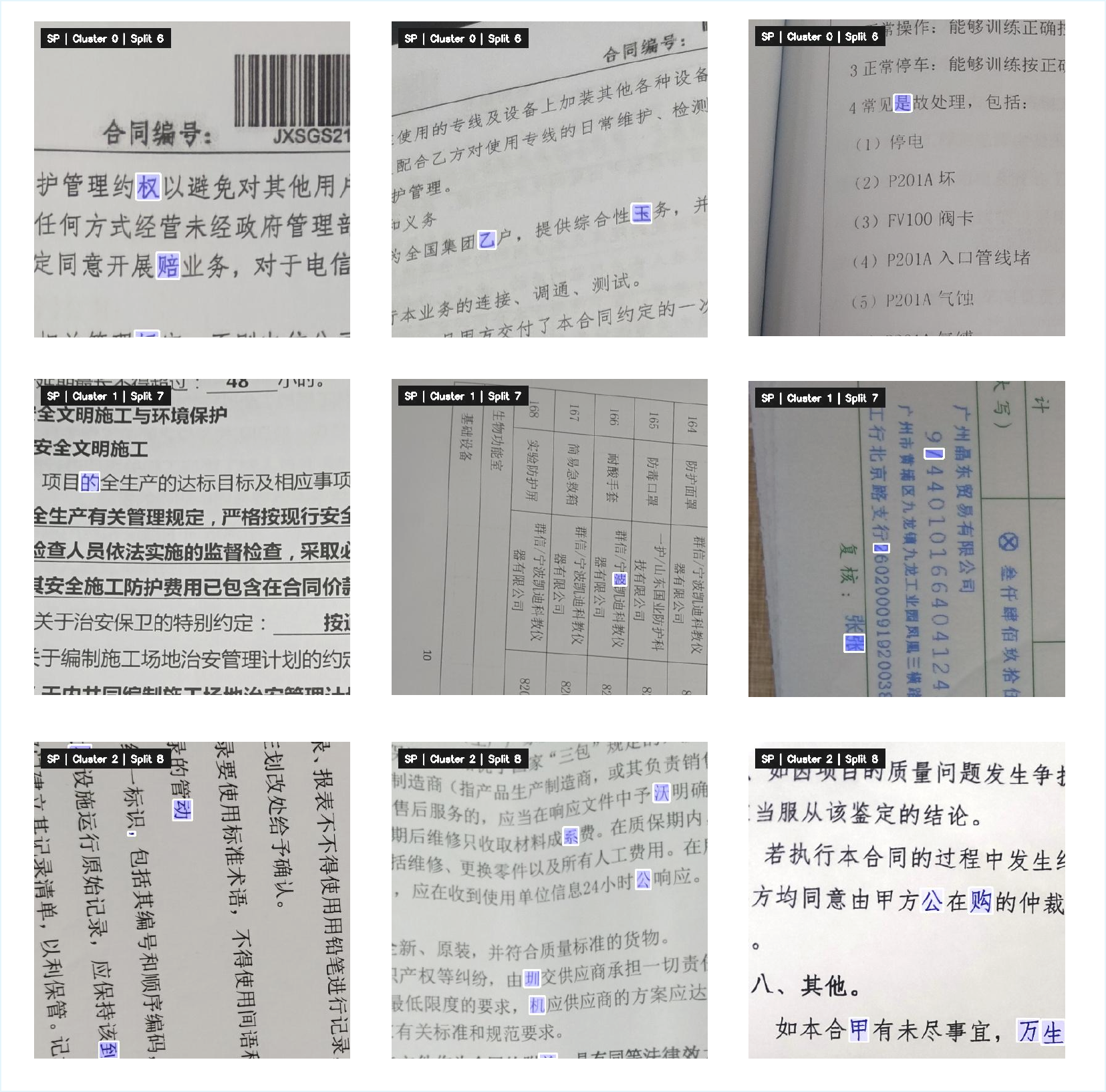}
        \caption{Splicing}
    \end{subfigure}
    \caption{Visualization of training samples and expert assignments. Each forgery synthesis type is further divided into three feature clusters, producing nine expert assignments in total. Within each subplot, each row shows samples from a single feature cluster.}
    \label{fig:train_samples}
\end{figure}

We visualize the samples for each subset of the training set in Fig.~\ref{fig:train_samples} and the corresponding expert assignments. The training set contains three forgery synthesis types: copy-move, splicing, and generation. Each type is further clustered into three feature clusters based on the query embedding of a trained RiSE-1 model, resulting in a total of nine expert assignments. The visualization shows that samples from different forgery types and feature clusters exhibit distinct visual characteristics, forgery traces, capturing conditions, and document semantics, which motivates the design of RiSE to learn specialized experts for each cluster.

\section{Limitation}

RiSE has several limitations. First, scaling the number of residual experts improves cross-domain F1 but requires evaluating all experts at inference. In the single-GPU measurement, RiSE-9 uses 21{,}691 MiB of memory and takes 0.3571 seconds per $512\times512$ image; this limits its use in resource-constrained settings. Early-exit expert pruning and residual-expert merging are promising directions for reducing this cost, but we have not evaluated them here. Second, LPC still leaves a gap to the oracle selector, and exact top-1 routing agreement is modest for RiSE-9. Third, RTM remains difficult because the source training data provide limited coverage of manual erasure, physical edits, and other handcrafted manipulations. Our added corruption experiments cover JPEG recompression and resizing; dedicated print-and-scan robustness evaluation remains future work.

\section{Broader Impacts}

The proliferation of document forgery poses significant risks to information integrity, financial security, and legal processes. By improving the robustness of document forgery detection across diverse manipulation types and domains, our work contributes to safeguarding against these threats. However, it is important to consider potential misuse of such detection technologies, such as infringing on privacy or being used for surveillance. We advocate for responsible deployment of forgery detection systems with appropriate safeguards to balance security benefits with ethical considerations.


\clearpage
\section*{NeurIPS Paper Checklist}

\begin{enumerate}

\item {\bf Claims}
    \item[] Question: Do the main claims made in the abstract and introduction accurately reflect the paper's contributions and scope?
    \item[] Answer: \answerYes{} 
    \item[] Justification: The claims match theoretical and experimental results.
    \item[] Guidelines:
    \begin{itemize}
        \item The answer \answerNA{} means that the abstract and introduction do not include the claims made in the paper.
        \item The abstract and/or introduction should clearly state the claims made, including the contributions made in the paper and important assumptions and limitations. A \answerNo{} or \answerNA{} answer to this question will not be perceived well by the reviewers. 
        \item The claims made should match theoretical and experimental results, and reflect how much the results can be expected to generalize to other settings. 
        \item It is fine to include aspirational goals as motivation as long as it is clear that these goals are not attained by the paper. 
    \end{itemize}

\item {\bf Limitations}
    \item[] Question: Does the paper discuss the limitations of the work performed by the authors?
    \item[] Answer: \answerYes{} 
    \item[] Justification: The paper includes a dedicated section discussing the limitations of the proposed approach.
    \item[] Guidelines:
    \begin{itemize}
        \item The answer \answerNA{} means that the paper has no limitation while the answer \answerNo{} means that the paper has limitations, but those are not discussed in the paper. 
        \item The authors are encouraged to create a separate ``Limitations'' section in their paper.
        \item The paper should point out any strong assumptions and how robust the results are to violations of these assumptions (e.g., independence assumptions, noiseless settings, model well-specification, asymptotic approximations only holding locally). The authors should reflect on how these assumptions might be violated in practice and what the implications would be.
        \item The authors should reflect on the scope of the claims made, e.g., if the approach was only tested on a few datasets or with a few runs. In general, empirical results often depend on implicit assumptions, which should be articulated.
        \item The authors should reflect on the factors that influence the performance of the approach. For example, a facial recognition algorithm may perform poorly when image resolution is low or images are taken in low lighting. Or a speech-to-text system might not be used reliably to provide closed captions for online lectures because it fails to handle technical jargon.
        \item The authors should discuss the computational efficiency of the proposed algorithms and how they scale with dataset size.
        \item If applicable, the authors should discuss possible limitations of their approach to address problems of privacy and fairness.
        \item While the authors might fear that complete honesty about limitations might be used by reviewers as grounds for rejection, a worse outcome might be that reviewers discover limitations that aren't acknowledged in the paper. The authors should use their best judgment and recognize that individual actions in favor of transparency play an important role in developing norms that preserve the integrity of the community. Reviewers will be specifically instructed to not penalize honesty concerning limitations.
    \end{itemize}

\item {\bf Theory assumptions and proofs}
    \item[] Question: For each theoretical result, does the paper provide the full set of assumptions and a complete (and correct) proof?
    \item[] Answer: \answerYes{} 
    \item[] Justification: The paper provides all necessary assumptions and complete proofs for the theoretical results.
    \item[] Guidelines:
    \begin{itemize}
        \item The answer \answerNA{} means that the paper does not include theoretical results. 
        \item All the theorems, formulas, and proofs in the paper should be numbered and cross-referenced.
        \item All assumptions should be clearly stated or referenced in the statement of any theorems.
        \item The proofs can either appear in the main paper or the supplemental material, but if they appear in the supplemental material, the authors are encouraged to provide a short proof sketch to provide intuition. 
        \item Inversely, any informal proof provided in the core of the paper should be complemented by formal proofs provided in appendix or supplemental material.
        \item Theorems and Lemmas that the proof relies upon should be properly referenced. 
    \end{itemize}

    \item {\bf Experimental result reproducibility}
    \item[] Question: Does the paper fully disclose all the information needed to reproduce the main experimental results of the paper to the extent that it affects the main claims and/or conclusions of the paper (regardless of whether the code and data are provided or not)?
    \item[] Answer: \answerYes{} 
    \item[] Justification: The paper provides all necessary details for reproducing the main experimental results, including dataset descriptions, model architectures, and training procedures.
    \item[] Guidelines:
    \begin{itemize}
        \item The answer \answerNA{} means that the paper does not include experiments.
        \item If the paper includes experiments, a \answerNo{} answer to this question will not be perceived well by the reviewers: Making the paper reproducible is important, regardless of whether the code and data are provided or not.
        \item If the contribution is a dataset and\slash or model, the authors should describe the steps taken to make their results reproducible or verifiable. 
        \item Depending on the contribution, reproducibility can be accomplished in various ways. For example, if the contribution is a novel architecture, describing the architecture fully might suffice, or if the contribution is a specific model and empirical evaluation, it may be necessary to either make it possible for others to replicate the model with the same dataset, or provide access to the model. In general. releasing code and data is often one good way to accomplish this, but reproducibility can also be provided via detailed instructions for how to replicate the results, access to a hosted model (e.g., in the case of a large language model), releasing of a model checkpoint, or other means that are appropriate to the research performed.
        \item While NeurIPS does not require releasing code, the conference does require all submissions to provide some reasonable avenue for reproducibility, which may depend on the nature of the contribution. For example
        \begin{enumerate}
            \item If the contribution is primarily a new algorithm, the paper should make it clear how to reproduce that algorithm.
            \item If the contribution is primarily a new model architecture, the paper should describe the architecture clearly and fully.
            \item If the contribution is a new model (e.g., a large language model), then there should either be a way to access this model for reproducing the results or a way to reproduce the model (e.g., with an open-source dataset or instructions for how to construct the dataset).
            \item We recognize that reproducibility may be tricky in some cases, in which case authors are welcome to describe the particular way they provide for reproducibility. In the case of closed-source models, it may be that access to the model is limited in some way (e.g., to registered users), but it should be possible for other researchers to have some path to reproducing or verifying the results.
        \end{enumerate}
    \end{itemize}

\item {\bf Open access to data and code}
    \item[] Question: Does the paper provide open access to the data and code, with sufficient instructions to faithfully reproduce the main experimental results, as described in supplemental material?
    \item[] Answer: \answerYes{} 
    \item[] Justification: The paper provides all necessary instructions and resources to reproduce the main experimental results, including access to code and model checkpoints.
    \item[] Guidelines:
    \begin{itemize}
        \item The answer \answerNA{} means that paper does not include experiments requiring code.
        \item Please see the NeurIPS code and data submission guidelines (\url{https://neurips.cc/public/guides/CodeSubmissionPolicy}) for more details.
        \item While we encourage the release of code and data, we understand that this might not be possible, so \answerNo{} is an acceptable answer. Papers cannot be rejected simply for not including code, unless this is central to the contribution (e.g., for a new open-source benchmark).
        \item The instructions should contain the exact command and environment needed to run to reproduce the results. See the NeurIPS code and data submission guidelines (\url{https://neurips.cc/public/guides/CodeSubmissionPolicy}) for more details.
        \item The authors should provide instructions on data access and preparation, including how to access the raw data, preprocessed data, intermediate data, and generated data, etc.
        \item The authors should provide scripts to reproduce all experimental results for the new proposed method and baselines. If only a subset of experiments are reproducible, they should state which ones are omitted from the script and why.
        \item At submission time, to preserve anonymity, the authors should release anonymized versions (if applicable).
        \item Providing as much information as possible in supplemental material (appended to the paper) is recommended, but including URLs to data and code is permitted.
    \end{itemize}

\item {\bf Experimental setting/details}
    \item[] Question: Does the paper specify all the training and test details (e.g., data splits, hyperparameters, how they were chosen, type of optimizer) necessary to understand the results?
    \item[] Answer: \answerYes{} 
    \item[] Justification: The paper provides all necessary details for reproducing the main experimental results, including dataset descriptions, model architectures, and training procedures.
    \item[] Guidelines:
    \begin{itemize}
        \item The answer \answerNA{} means that the paper does not include experiments.
        \item The experimental setting should be presented in the core of the paper to a level of detail that is necessary to appreciate the results and make sense of them.
        \item The full details can be provided either with the code, in appendix, or as supplemental material.
    \end{itemize}

\item {\bf Experiment statistical significance}
    \item[] Question: Does the paper report error bars suitably and correctly defined or other appropriate information about the statistical significance of the experiments?
    \item[] Answer: \answerYes{} 
    \item[] Justification: The paper evaluates the proposed method on large-scale test sets.
    \item[] Guidelines:
    \begin{itemize}
        \item The answer \answerNA{} means that the paper does not include experiments.
        \item The authors should answer \answerYes{} if the results are accompanied by error bars, confidence intervals, or statistical significance tests, at least for the experiments that support the main claims of the paper.
        \item The factors of variability that the error bars are capturing should be clearly stated (for example, train/test split, initialization, random drawing of some parameter, or overall run with given experimental conditions).
        \item The method for calculating the error bars should be explained (closed form formula, call to a library function, bootstrap, etc.)
        \item The assumptions made should be given (e.g., Normally distributed errors).
        \item It should be clear whether the error bar is the standard deviation or the standard error of the mean.
        \item It is OK to report 1-sigma error bars, but one should state it. The authors should preferably report a 2-sigma error bar than state that they have a 96\% CI, if the hypothesis of Normality of errors is not verified.
        \item For asymmetric distributions, the authors should be careful not to show in tables or figures symmetric error bars that would yield results that are out of range (e.g., negative error rates).
        \item If error bars are reported in tables or plots, the authors should explain in the text how they were calculated and reference the corresponding figures or tables in the text.
    \end{itemize}

\item {\bf Experiments compute resources}
    \item[] Question: For each experiment, does the paper provide sufficient information on the computer resources (type of compute workers, memory, time of execution) needed to reproduce the experiments?
    \item[] Answer: \answerYes{} 
    \item[] Justification: The paper provides the hardware specifications and computational requirements for reproducing the experiments.
    \item[] Guidelines:
    \begin{itemize}
        \item The answer \answerNA{} means that the paper does not include experiments.
        \item The paper should indicate the type of compute workers CPU or GPU, internal cluster, or cloud provider, including relevant memory and storage.
        \item The paper should provide the amount of compute required for each of the individual experimental runs as well as estimate the total compute. 
        \item The paper should disclose whether the full research project required more compute than the experiments reported in the paper (e.g., preliminary or failed experiments that didn't make it into the paper). 
    \end{itemize}
    
\item {\bf Code of ethics}
    \item[] Question: Does the research conducted in the paper conform, in every respect, with the NeurIPS Code of Ethics \url{https://neurips.cc/public/EthicsGuidelines}?
    \item[] Answer: \answerYes{} 
    \item[] Justification: The research conducted in the paper adheres to the NeurIPS Code of Ethics.
    \item[] Guidelines:
    \begin{itemize}
        \item The answer \answerNA{} means that the authors have not reviewed the NeurIPS Code of Ethics.
        \item If the authors answer \answerNo, they should explain the special circumstances that require a deviation from the Code of Ethics.
        \item The authors should make sure to preserve anonymity (e.g., if there is a special consideration due to laws or regulations in their jurisdiction).
    \end{itemize}

\item {\bf Broader impacts}
    \item[] Question: Does the paper discuss both potential positive societal impacts and negative societal impacts of the work performed?
    \item[] Answer: \answerYes{} 
    \item[] Justification: The paper discusses both potential positive and negative societal impacts of the work performed.
    \item[] Guidelines:
    \begin{itemize}
        \item The answer \answerNA{} means that there is no societal impact of the work performed.
        \item If the authors answer \answerNA{} or \answerNo, they should explain why their work has no societal impact or why the paper does not address societal impact.
        \item Examples of negative societal impacts include potential malicious or unintended uses (e.g., disinformation, generating fake profiles, surveillance), fairness considerations (e.g., deployment of technologies that could make decisions that unfairly impact specific groups), privacy considerations, and security considerations.
        \item The conference expects that many papers will be foundational research and not tied to particular applications, let alone deployments. However, if there is a direct path to any negative applications, the authors should point it out. For example, it is legitimate to point out that an improvement in the quality of generative models could be used to generate Deepfakes for disinformation. On the other hand, it is not needed to point out that a generic algorithm for optimizing neural networks could enable people to train models that generate Deepfakes faster.
        \item The authors should consider possible harms that could arise when the technology is being used as intended and functioning correctly, harms that could arise when the technology is being used as intended but gives incorrect results, and harms following from (intentional or unintentional) misuse of the technology.
        \item If there are negative societal impacts, the authors could also discuss possible mitigation strategies (e.g., gated release of models, providing defenses in addition to attacks, mechanisms for monitoring misuse, mechanisms to monitor how a system learns from feedback over time, improving the efficiency and accessibility of ML).
    \end{itemize}
    
\item {\bf Safeguards}
    \item[] Question: Does the paper describe safeguards that have been put in place for responsible release of data or models that have a high risk for misuse (e.g., pre-trained language models, image generators, or scraped datasets)?
    \item[] Answer: \answerNA{} 
    \item[] Justification: The paper poses no such risks.
    \item[] Guidelines:
    \begin{itemize}
        \item The answer \answerNA{} means that the paper poses no such risks.
        \item Released models that have a high risk for misuse or dual-use should be released with necessary safeguards to allow for controlled use of the model, for example by requiring that users adhere to usage guidelines or restrictions to access the model or implementing safety filters. 
        \item Datasets that have been scraped from the Internet could pose safety risks. The authors should describe how they avoided releasing unsafe images.
        \item We recognize that providing effective safeguards is challenging, and many papers do not require this, but we encourage authors to take this into account and make a best faith effort.
    \end{itemize}

\item {\bf Licenses for existing assets}
    \item[] Question: Are the creators or original owners of assets (e.g., code, data, models), used in the paper, properly credited and are the license and terms of use explicitly mentioned and properly respected?
    \item[] Answer: \answerYes{} 
    \item[] Justification: The paper properly credits all creators and original owners of assets used, and the licenses and terms of use are explicitly mentioned and respected.
    \item[] Guidelines:
    \begin{itemize}
        \item The answer \answerNA{} means that the paper does not use existing assets.
        \item The authors should cite the original paper that produced the code package or dataset.
        \item The authors should state which version of the asset is used and, if possible, include a URL.
        \item The name of the license (e.g., CC-BY 4.0) should be included for each asset.
        \item For scraped data from a particular source (e.g., website), the copyright and terms of service of that source should be provided.
        \item If assets are released, the license, copyright information, and terms of use in the package should be provided. For popular datasets, \url{paperswithcode.com/datasets} has curated licenses for some datasets. Their licensing guide can help determine the license of a dataset.
        \item For existing datasets that are re-packaged, both the original license and the license of the derived asset (if it has changed) should be provided.
        \item If this information is not available online, the authors are encouraged to reach out to the asset's creators.
    \end{itemize}

\item {\bf New assets}
    \item[] Question: Are new assets introduced in the paper well documented and is the documentation provided alongside the assets?
    \item[] Answer: \answerYes{} 
    \item[] Justification: The paper provides detailed documentation for all new assets introduced, including datasets, code, and models.
    \item[] Guidelines:
    \begin{itemize}
        \item The answer \answerNA{} means that the paper does not release new assets.
        \item Researchers should communicate the details of the dataset\slash code\slash model as part of their submissions via structured templates. This includes details about training, license, limitations, etc. 
        \item The paper should discuss whether and how consent was obtained from people whose asset is used.
        \item At submission time, remember to anonymize your assets (if applicable). You can either create an anonymized URL or include an anonymized zip file.
    \end{itemize}

\item {\bf Crowdsourcing and research with human subjects}
    \item[] Question: For crowdsourcing experiments and research with human subjects, does the paper include the full text of instructions given to participants and screenshots, if applicable, as well as details about compensation (if any)? 
    \item[] Answer: \answerNA{} 
    \item[] Justification: The paper does not involve crowdsourcing nor research with human subjects.
    \item[] Guidelines:
    \begin{itemize}
        \item The answer \answerNA{} means that the paper does not involve crowdsourcing nor research with human subjects.
        \item Including this information in the supplemental material is fine, but if the main contribution of the paper involves human subjects, then as much detail as possible should be included in the main paper. 
        \item According to the NeurIPS Code of Ethics, workers involved in data collection, curation, or other labor should be paid at least the minimum wage in the country of the data collector. 
    \end{itemize}

\item {\bf Institutional review board (IRB) approvals or equivalent for research with human subjects}
    \item[] Question: Does the paper describe potential risks incurred by study participants, whether such risks were disclosed to the subjects, and whether Institutional Review Board (IRB) approvals (or an equivalent approval/review based on the requirements of your country or institution) were obtained?
    \item[] Answer: \answerNA{} 
    \item[] Justification: The paper does not involve crowdsourcing nor research with human subjects.
    \item[] Guidelines:
    \begin{itemize}
        \item The answer \answerNA{} means that the paper does not involve crowdsourcing nor research with human subjects.
        \item Depending on the country in which research is conducted, IRB approval (or equivalent) may be required for any human subjects research. If you obtained IRB approval, you should clearly state this in the paper. 
        \item We recognize that the procedures for this may vary significantly between institutions and locations, and we expect authors to adhere to the NeurIPS Code of Ethics and the guidelines for their institution. 
        \item For initial submissions, do not include any information that would break anonymity (if applicable), such as the institution conducting the review.
    \end{itemize}

\item {\bf Declaration of LLM usage}
    \item[] Question: Does the paper describe the usage of LLMs if it is an important, original, or non-standard component of the core methods in this research? Note that if the LLM is used only for writing, editing, or formatting purposes and does \emph{not} impact the core methodology, scientific rigor, or originality of the research, declaration is not required.
    \item[] Answer: \answerNA{} 
    \item[] Justification: The paper does not use LLMs as an important, original, or non-standard component.
    \item[] Guidelines:
    \begin{itemize}
        \item The answer \answerNA{} means that the core method development in this research does not involve LLMs as any important, original, or non-standard components.
        \item Please refer to our LLM policy in the NeurIPS handbook for what should or should not be described.
    \end{itemize}

\end{enumerate}

\end{document}